\documentclass{article}
\usepackage{iclr2027_conference,times}

\usepackage{amsmath,amsfonts,bm}

\def\eqref#1{equation~\ref{#1}}

\def\1{\bm{1}}

\DeclareMathAlphabet{\mathsfit}{\encodingdefault}{\sfdefault}{m}{sl}
\SetMathAlphabet{\mathsfit}{bold}{\encodingdefault}{\sfdefault}{bx}{n}

\usepackage{amsmath,amssymb}
\usepackage{graphicx}
\usepackage{booktabs}
\usepackage{multirow}
\usepackage{algorithm}
\usepackage{algpseudocode}

\usepackage{xcolor}
\usepackage{url}
\usepackage{hyperref}
\usepackage[capitalize,noabbrev]{cleveref}

\graphicspath{{figures/}}

\title{AVQ-Attention-2:\ Adaptive Vector-Quantized Attention with Exact Local Blocks and Linear-Cost Training}

\author{Winfried van den Dool\\QUVA Lab \& AMLab, University of Amsterdam\\\texttt{w.v.s.o.vandendool@uva.nl}
\AND Patrick Forr\'{e}\\AI4Science Lab \& KdV Institute, University of Amsterdam
\And Amir Habibian\\Qualcomm AI Research, Amsterdam
\AND Yuki M.~Asano\\FunAI Lab, University of Technology Nuremberg
\And Max Welling\\AMLab, University of Amsterdam}
\iclrfinalcopy

\begin{document}

\maketitle

\begin{abstract}
Attention costs $\mathcal{O}(N^2)$ in the number of tokens. Vector-quantized attention reduces this to $\mathcal{O}(NM)$ by clustering the $N$ keys, attending to the $M$ codewords instead, and its adaptive form, AVQ-attention, concentrates codebook resolution where a query's attention mass actually falls. We present AVQ-Attention-2, a better design that merges it with exact local attention and attends to the mean of each cluster's keys rather than to a codeword. We show three improvements: consistently better quality at comparable cost; training that now also costs $\mathcal{O}(NM)$ rather than $\mathcal{O}(N^2)$; and an approximation accurate enough that the layer can be dropped into a pretrained transformer without fine-tuning. Across a variety of sequence lengths the result is competitive not merely with other approximate attention but with exact dense attention, running faster than FlashAttention at little to no loss in quality.
\end{abstract}

\section{Introduction}
\label{sec:introduction}

Attention compares every token with every other, so its cost grows quadratically with sequence length. FlashAttention \citep{dao2022flashattention,dao2024flashattention2} removes the memory bottleneck of that computation but not its arithmetic: the quadratic term counts pairs of tokens, and an exact method has to look at all of them. Cheaper attention therefore has to decide what not to pay for, which leaves two alternatives: cut the pairs deemed less important and keep the rest exact, or keep every pair and approximate it.

In the first camp, local attention methods like windowed and neighbourhood attention \citep{liu2021swin,hassani2023nat} restrict every query to a nearby subset of keys and evaluate those exactly. In doing so they lose global information, and at high resolutions the neighbourhoods have to grow in step to keep proportional coverage. Vector-quantized attention instead sorts the $N$ keys into $M$ clusters and attends over the clusters \citep{lingletransformer}, so every key still contributes, through a coarser stand-in; AVQ-attention \citep{vandendool2026avq} makes that resolution adaptive, spending a per-tile refinement budget on whichever key clusters receive the most attention mass.

However, attention mass is often concentrated near the query. Local sparse methods build that prior in directly; AVQ-attention has to rediscover it from the data, and then spends part of its budget approximating keys that a local method computes exactly, and faster. Seen that way the two camps are best used as complements: compute each query's own neighbourhood exactly, and let adaptive approximation cover the rest. The benefit does not come only from the exact local attention, but also from the adaptive capacity it frees for global structure.

We propose AVQ-Attention-2, an adaptive vector-quantized attention layer that does both at once. Every query attends exactly over its own neighbourhood and through cluster representatives to everything beyond it. Critically, we develop an efficient kernel implementation that allows this to happen within a single attention pass, rather than as two mechanisms run side by side, all the while never materializing the attention matrix, as in FlashAttention. Careful bookkeeping in the kernel code allows focusing the adaptive part entirely outside that neighbourhood, effectively using the compute for better global attention at the same time.

On top of combining AVQ-attention with exact local attention, we propose improvements to each. For the local attention, we propose a hardware-aware method that computes optimally shaped, compact neighbourhoods applied once before training. We show that such hardware-awareness and principled neighbourhood selection is Pareto-optimal compared to existing local attention frameworks (\cref{sec:exp_local}). For AVQ-attention the codebook vector itself is actually not required at all: only the fact that keys are clustered in the first place. We can thus use the key cluster's own mean instead of the fixed codeword to create a better approximation. As the mean of keys is a differentiable function of those keys, the layer can be trained through it directly. The codebook is kept purely as a routing mechanism, updated in the background by an exponential moving average of the clusters it produces, and never has to interact with the model's gradients. Furthermore, we improve the speed of the quantization pass by first sorting the keys by their cluster assignment, which lets each cluster's quantities be written directly rather than accumulated through contention.

As the effectiveness of linear-time attention methods depends a lot on sequence length, we evaluate our method to cover a range of different $N$: image classification on ImageNet-1k \citep{deng2009imagenet} ($N = 785$), semantic segmentation on ADE20K \citep{zhou2017ade20k} ($N = 901$) and a V-JEPA~2 video encoder \citep{assran2025vjepa2} on Diving-48 \citep{li2018diving48} ($N = 9216$ and $18432$). While the biggest benefits come at large $N$, the smaller settings can be the less forgiving test, since a method cannot hide there behind better asymptotic behaviour.

Our contributions are:
\begin{itemize}
\item Improved design of both hardware-aware local attention and adaptive vector-quantized attention separately, as well as an implementation that combines both.
\item Custom Triton \citep{tillet2019triton} kernels that fuse that combination into a single attention pass: the exact local keys and the cluster representatives share the same online softmax and normalizer, so that no intermediate output has to be materialized.
\item We show that the result is an attention layer that runs in linear time for both training and inference, already competitive with FlashAttention at sequence lengths as short as $N=785$.
\end{itemize}

\section{Related Work}
\label{sec:related_work}

\paragraph{Local and sparse attention.}
Restricting attention to a neighbourhood around each query is among the most effective efficient-attention families in practice, and underpins the standard efficient backbones in vision as well as long-document models \citep{weissenborn2020scaling,liu2021swin,hassani2023nat,beltagy2020longformerlongdocumenttransformer,child2019generatinglongsequencessparse}. The Routing Transformer \citep{roy2021efficient} combines local attention heads with sparse global ones, clustering key--query space and letting each query attend only to the keys sharing its centroid, dropping the remaining pairs. Another idea is to shape the sparse pattern so a fused kernel can exploit it: MoBA \citep{lu2025moba} and NSA \citep{yuan2025native} select whole blocks rather than individual keys, since GPUs read contiguous blocks far faster than scattered indices. We apply the same reasoning to local attention: the tile width is the granularity the attention kernel already works at, so we take it as given and spend our freedom on the shape and placement of the tiles instead. Others have varied tile shape as well, but more elaborately, adapting it to the input on every forward pass \citep{zhang2022vsa,yu2022boat}; we optimise one arrangement against the hardware once, before training, which makes it free at run time.

\paragraph{Attention over group representatives.}
A second line of work lets a query attend to a representative of a group of keys rather than to every key in it. Fast Multipole Attention \citep{kang2025fma} and H-Transformer-1D \citep{zhu2021htransformer} summarise distant groups at several scales, the first by a learned weighted sum of each group's keys and values, the second by averaging and over coarsened queries as well. MRA attention \citep{zeng2022mra} estimates the attention matrix from block means before recomputing the highest-scoring blocks exactly, and Token Merging \citep{bolya2022tokenmerging} merges similar tokens, weighting each merged term by the number it stands for. SVG-EAR \citep{zhou2026svgear} clusters keys and queries on every call and decides per cluster pair between a centroid approximation and exact attention. Agent Attention \citep{han2024agent} and mixture-of-top-$k$ attention \citep{wen2026mita} attend through average-pooled queries, each carrying its own attention over all keys as its value; mixture-of-top-$k$ attention additionally routes each query to the keys its pooled query scores highest, attended exactly. These methods differ mainly in the construction they use to remove the quadratic term. However, FlashAttention \citep{dao2022flashattention,dao2024flashattention2} has moved the goalposts: memory traffic now counts alongside arithmetic, and an efficiency that rests on avoiding quadratic flops alone will be modest and appear only at very long sequences. The cost of building and reading the representatives thus becomes decisive: forming the value of a pooled query takes an attention over all the keys \citep{han2024agent,wen2026mita}, clustering on every call takes repeated passes over the keys \citep{zhou2026svgear}, and selecting from a coarse estimate takes gathers at data-dependent indices \citep{zeng2022mra,wen2026mita}.

\paragraph{Vector-quantized attention.}
A specific implementation of group-representative attention is to use vector quantization (VQ). Transformer-VQ \citep{lingletransformer} quantizes each key to a learned codebook and attends over the codewords instead of the keys. As such it exploits the advantage of clustering, having groups formed from the keys themselves rather than fixed, but pays for it only during training, when the codebook is learned. Inference for VQ is cheap, with an assignment pass that can be tiled like FlashAttention, never materialising the key--codeword distances. AVQ-attention \citep{vandendool2026avq} builds on this by adding a hierarchical codebook whose refinement is chosen per query tile from the attention mass each cluster receives. VQ-attention has also been applied to causal language models, with the codebook maintained online rather than learned \citep{alonso2026ovq}, and to graphs, over centroids built in input-feature space \citep{kong2023goat}. We build on AVQ-attention's hierarchy---without losing its efficient tiled implementation---by bringing each query's own exact neighbourhood into the same online softmax as the cluster representatives. Furthermore, using cluster means rather than codewords effectively corresponds to one k-means update on top of the assignments, which can be viewed as a hybrid between clustering and quantization: centroids fitted to the keys at inference time, with the majority of the clustering work already done while training.

\paragraph{Linear attention and state-space models.}
Kernelized attention \citep{katharopoulos2020linear,choromanski2021performer} replaces the softmax and low-rank attention \citep{wang2020linformer} projects the keys and values to a shorter sequence, while state-space models \citep{gu2022s4,gu2024mamba} replace attention altogether, mixing sequences in linear time at the cost of content-based lookup. Scatterbrain \citep{chen2021scatterbrain} combines the two: a random-feature approximation of the whole attention matrix, corrected on the entries that locality-sensitive hashing selects so that it is exact there. We keep standard softmax attention and reduce the number of distinct objects each query compares against.

\paragraph{Quantization of the attention computation.}
A separate line reduces the bit width of the attention computation itself: SageAttention \citep{zhang2025sage} runs the products in 8 bits, and the key--value cache can be stored at fewer bits to shrink the cache during decoding \citep{li2025commvq,hooper2024kvquant}. That axis is orthogonal to ours, and composable with it: those methods keep $N$ interactions and make each one cheaper, where we reduce the number of interactions, but the compute inside our layer can be quantized to lower bitwidth the same way.

\section{Preliminaries}
\label{sec:preliminaries}

\subsection{Vector-Quantized Attention}
\label{sec:prelim_vq}

Attention computes $Y_i = \sum_j \exp(Q_i K_j^{\top}) V_j \,/\, \sum_l \exp(Q_i K_l^{\top})$ over $N$ tokens, at $\mathcal{O}(N^2 D)$ cost. Vector-quantized attention replaces each key by a representative drawn from a codebook $\{C_a\}_{a=1}^{M}$ with $M \ll N$, assigning $a(K_j) = \arg\min_a \|C_a - K_j\|^2$. Keys sharing a representative then share a logit, so the sums can be regrouped per code. With
\begin{equation}
\label{eq:aggregates}
n_a = |\{j : a(K_j) = a\}|, \qquad \bar V_a = \sum_{j:\,a(K_j)=a} V_j ,
\end{equation}
attention becomes a softmax over $M$ codes rather than $N$ keys,
\begin{equation}
\label{eq:vqattention}
Y_i \;\approx\; \frac{1}{\bar Z_i} \sum_{a=1}^{M} \exp(Q_i C_a^{\top})\, \bar V_a ,
\qquad
\bar Z_i \;=\; \sum_{a=1}^{M} \exp(Q_i C_a^{\top})\, n_a ,
\end{equation}

\subsection{Adaptive VQ-Attention}
\label{sec:prelim_avq}

AVQ-attention \citep{vandendool2026avq} makes the codebook hierarchical and its resolution in key space input-dependent. A codebook of $M_0$ parents, each with $\mathcal{C}$ children, is traversed per key: a key is assigned to its nearest parent and then, if a child is closer, to that child. Attention proceeds in two stages of one FlashAttention-style pass \citep{dao2022flashattention}.\footnote{The numerator and the denominator $\bar Z_i$ of \cref{eq:vqattention} are carried as running sums per query under an online softmax; the attention matrix is never materialized.} The first stage computes \cref{eq:vqattention} over the $M_0$ parents and, from the running denominator, an importance score per code and query tile $I$,
\begin{equation}
\label{eq:importance}
w_j(I) \;=\; \sum_{i \in I} \frac{A_{ij}\, n_j}{\bar Z_i}, \qquad A_{ij} = \exp(Q_i C_j^{\top}),
\end{equation}
i.e.\ the share of tile $I$'s attention mass that code $j$ receives. The second stage \emph{refines} the $\mathcal{P}$ parents with the largest $w_j(I)$: attention is also computed over their children. Writing $S_{ip} = Q_i C_p^{\top}$ for the parent's logit and $S_{ic}$ for a child's, the keys that were closer to a child than to the parent have their contribution to the running numerator and denominator replaced by that child's, by accumulating the \emph{correcting attention}
\begin{equation}
\label{eq:corr}
\Delta A_{ic} \;=\; \exp(S_{ic}) - \exp(S_{ip})
\end{equation}
against the child aggregates $\bar V_c$ and $n_c$: the $\exp(S_{ic})$ puts a key in at its child code, the $\exp(S_{ip})$ takes it out at its parent code, in the same dot product. To avoid having to recompute or store the parent logit, AVQ-attention constrains each parent to be the mean of its children, $C_p = \frac{1}{\mathcal{C}}\sum_c C_{p,c}$, so $S_{ip} = \frac{1}{\mathcal{C}}\sum_c S_{ic}$ is a reduction over child logits already in registers.

While inference costs $\mathcal{O}((M_0 + \mathcal{P}\mathcal{C}) N D)$, training is slightly more complicated. The assignment $a(K_j)$ is a discrete choice, so no gradient reaches the keys through it, and AVQ-attention trains with a straight-through estimator that attends over all $N$ quantized keys, making the forward pass that the gradient flows through quadratic in $N$. 

Finally, refinement is done per query tile, so different regions of the input refine different parts of the codebook. To make tiles spatially compact, tokens are reordered along a Gilbert space-filling curve \citep{cerveny2018gilbert}.

\section{Method}
\label{sec:method}

\subsection{Attending to Cluster Means}
\label{sec:method_keymeans}
Because the regrouping in \cref{eq:vqattention} holds for any vector shared by the keys of a cluster, we can use the mean of the keys actually assigned to it,
\begin{equation}
\label{eq:keymeans}
\hat K_j \;=\; \bar K_{m(j)}, \qquad \bar K_m \;=\; \frac{1}{n_m}\sum_{j:\,m(j)=m} K_j ,
\end{equation}
computed per input rather than stored. The codebook, trained using online k-means as in \cite{vandendool2026avq}, keeps its role in deciding \emph{which} keys form a cluster, but the actual $\bar K_m$ is what is used in attention. Naturally, as the minimizer of $\sum_{j \in m}\|K_j - x\|^2$, this reduces quantization error. Because it is also based directly on current keys, attention over means tends to exact attention as refinement grows.

\paragraph{Training becomes exact and linear in $N$.}
Unlike a codeword, $\bar K_m$ is a differentiable function of the keys: $\partial \bar K_{m,d} / \partial K_{j',d} = \mathbf{1}[m(j') = m]/n_m$. Write $p_{im} = \exp(Q_i \bar K_m^{\top})\, n_m / \bar Z_i$ for the weight cluster $m$ receives from query $i$ in \cref{eq:vqattention}, $O_i = \sum_m p_{im} \bar V_m / n_m$ for the layer output and $dO_i = \partial L / \partial O_i$ for the gradient arriving at it. The chain rule collapses the per-key gradient onto its cluster,
\begin{equation}
\label{eq:collapse}
\frac{\partial L}{\partial K_{j',d}}
\;=\; \frac{1}{n_m}\sum_{i} p_{im}\,\big\langle dO_i,\ \bar V_m/n_m - O_i \big\rangle\, Q_{i,d}, \qquad m = m(j'),
\end{equation}
derived in \cref{app:derivations}. The per-key factor $V_j$ that made the straight-through path quadratic has averaged into $\bar V_m$ by linearity; every quantity on the right is per cluster or per query, so the backward pass costs $\mathcal{O}(MND)$, matching the forward.

\paragraph{The hierarchy, without the constraint.}
Refinement never recomputes a coarse representative's logit; it rebuilds it from the finer logits that replace it, which is what keeps it cheap. AVQ-attention can do so because each parent is constrained to be the mean of its children. A cluster mean satisfies no such constraint: $\bar K_p$ need not equal $\frac{1}{\mathcal{C}}\sum_c \bar K_{p,c}$. A count-weighted mean would solve this, but only if no key stays behind at the parent, so that $n_p = \sum_c n_c$. In refinement, contrary to AVQ-attention, we therefore let \textit{every key} descend to its nearest child, and can then write
\begin{equation}
\label{eq:weighted_parent}
\bar K_p \;=\; \sum_{c=1}^{\mathcal{C}} \frac{n_c}{n_p}\, \bar K_{p,c},
\qquad\text{hence}\qquad
S_{ip} \;=\; \sum_{c=1}^{\mathcal{C}} \frac{n_c}{n_p}\, S_{ic}.
\end{equation}
The parent logit is then again a reduction over child logits already in registers. As an additional benefit the constraint has no remaining job: children are free to move to optimal positions as a plain clustering of their parent's cell.

\subsection{Combining Exact Local and AVQ-Attention}
\label{sec:method_localexact}

A query's nearest keys are where most of its attention mass sits. We therefore let each query tile attend to its own $\text{Block}_N$ keys exactly and to cluster means everywhere else. However, naively applying the two together runs into a clear issue: a key the tile attends to exactly is still a member of its cluster, so it would contribute once as itself and once through $\bar V_m$.

For simplicity we first consider the denominator, writing $n_m^{\,I} = |\{j \in I : m(j) = m\}|$ for the number of cluster $m$'s keys that lie in tile $I$. Every cluster's count separates into the keys the tile does not hold and the keys it does,
\begin{align}
\label{eq:split}
\bar Z_I = \sum_{m} \exp(Q_I \bar K_m^{\top})\, n_m
&= \sum_{m} \exp(Q_I \bar K_m^{\top}) \big(n_m - n_m^{\,I}\big)
+ \sum_{m} \exp(Q_I \bar K_m^{\top})\, n_m^{\,I} \nonumber \\
&= \sum_{m} \exp(Q_I \bar K_m^{\top}) \big(n_m - n_m^{\,I}\big)
+ \sum_{j \in I} \exp(Q_I \bar K_{m(j)}^{\top}) .
\end{align}
The second sum is exactly the tile's own keys, each still counted through its cluster mean. Replacing it by the same sum over the keys themselves, every cluster is left counting only its keys outside the tile,
\begin{equation}
\label{eq:denomlocal}
\bar Z_I = \sum_{m} \exp(Q_I \bar K_m^{\top}) \big(n_m - n_m^{\,I}\big) \;+\; \sum_{j \in I} \exp(Q_I K_j^{\top}) .
\end{equation}
This is what the tile computes, and it costs little: $n_m^{\,I}$ is a histogram of $M$ integers, subtracted before the pass begins.

The same split does not transfer to the numerator. The tempting analogue, a per-tile value sum $\bar V_m^{\,I}$, is a $D$-wide aggregate per cluster rather than a single integer, so forming it would mean a scatter inside every tile. What's more, $m(j)$ is not yet known when the cluster terms are accumulated, since which code a local key is represented by depends on whether this tile refines its parent, decided only after the parent pass. \footnote{One may end up pre-emptively removing $\bar V_m^{\,I}$ for a parent code $m$ to later find that in fact parent $m$ is replaced by its children.} The numerator instead keeps the full $\bar V_m$, and folds the correction directly into the epilogue that computes exact local attention, once each key has a final code:
\begin{equation}
\label{eq:localexact}
\sum_{m} \exp(Q_I \bar K_m^{\top})\, \bar V_m
\;+\; \sum_{j \in I} \Big[\exp(Q_I K_j^{\top}) - \exp(Q_I \bar K_{m(j)}^{\top})\Big] V_j ,
\end{equation}
which resolves back into the aggregate form the tile cannot compute,
\begin{equation}
\label{eq:localexact_agg}
\sum_{m} \exp(Q_I \bar K_m^{\top}) \big(\bar V_m - \bar V_m^{\,I}\big)
\;+\; \sum_{j \in I} \exp(Q_I K_j^{\top})\, V_j .
\end{equation}

\begin{figure}[t]
\begin{center}
\includegraphics[width=\linewidth]{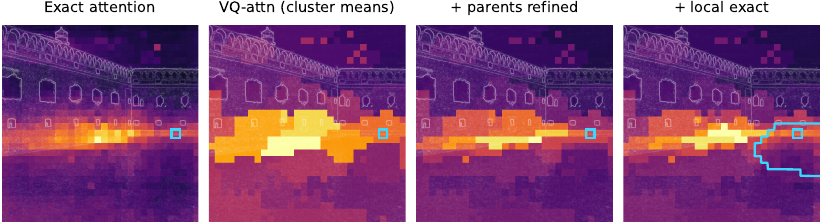}
\end{center}
\caption{Attention received by one query (cyan), computed four ways. Exact local attention is meant to fix
the query's own neighbourhood, but it also sharpens the rest of the map: once the tile's own keys are
computed exactly they no longer attract refinement, so the budget is spent on the rest of the image. The
share of the selected parents' keys lying inside the tile falls from $0.33$ to $0.06$.}
\label{fig:attention_maps}
\end{figure}

We thus perform a FlashAttention-style pass by first tiling over cluster means and then finishing with a key tile. The online softmax does not care that the numerator and the denominator are accumulated differently, only that the aggregate it ends with is correct. To exploit the synergy with AVQ-attention, however, it \textit{is} important that the corrected counts $n_m - n_m^{\,I}$ are used directly: the importance score of \cref{eq:importance} is derived from them, so the parents selected for refinement are more likely to be picked from global context, rather than overlapping with the local tile that is already covered exactly (\cref{fig:attention_maps}).

A second synergy is particular to the cluster-mean formulation. \Cref{eq:collapse} collapses each key's gradient onto its cluster, which is exactly what makes training linear in $N$; the price is that a key no longer has a training signal of its own, only its share of its cluster's, and keys are updated as members of a cluster rather than as themselves. Exact local attention gives that back. A key attended to exactly by the $\text{Block}_N$ queries of its own tile enters the loss as itself, and is moved by what it did rather than by what its cluster did. The two terms then carry complementary signal: the cluster term settles where a cluster sits, the exact term separates the keys inside it.

\subsection{Choosing the Tiles}
\label{sec:method_tiling}

The kernel processes queries in tiles of $\text{Block}_N$, and local attention computes each tile's own keys exactly: the tile \textit{is} the local neighbourhood. By folding a permutation of $N$ into the kernels, however, which tokens \textit{share} a tile can still be freely chosen. Compactness helps both the adaptive refinement --- nearby queries sharing a decision which parents to refine --- and the exact local attention, making it the obvious thing to optimise for. However, consecutive layers may also benefit from shifted tiles, so that a token at the boundary of its tile in one layer sits near the centre of one in the next.

\begin{figure}[t]
\begin{center}
\includegraphics[width=\linewidth]{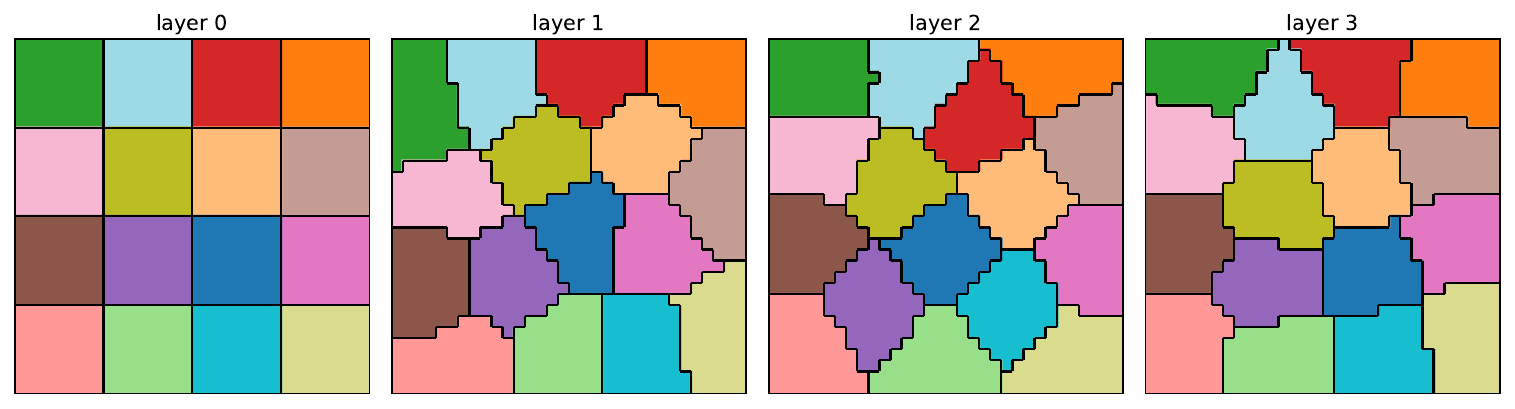}
\end{center}
\caption{Four consecutive layers of a $32 \times 32$ token grid at $\text{Block}_N = 64$. Each layer
places its boundaries where the previous layers did not have them.}
\label{fig:balanced_tiling}
\end{figure}

Write $x_i \in \mathbb{R}^{d}$ for the position of token $i$, $\pi$ for the map from tokens to tiles of exactly $\text{Block}_N$ tokens each, and $c_t$ for the centre of tile $t$. For each layer $l$ we solve (\cref{app:tiling})
\begin{equation}
\label{eq:balanced}
\min_{\pi,\,c} \; \sum_{i=1}^{N} w^{(l)}_i \lVert x_i - c_{\pi(i)} \rVert^{2}
\qquad \text{subject to} \qquad
\bigl\lvert \pi^{-1}(t) \bigr\rvert = \text{Block}_N \quad \text{for every } t ,
\end{equation}
where the weight $w^{(l)}_i$ lets one token ask to sit closer to a centre than another, if earlier layers placed it badly. How badly is exactly $\lVert x_i - c_{\pi(i)} \rVert^{2}$, so we accumulate it over layers with an exponential moving average,
\begin{equation}
\label{eq:layerweights}
a^{(l)}_i \;=\; \gamma\, a^{(l-1)}_i \;+\; \bigl\lVert x_i - c^{(l-1)}_{\pi_{l-1}(i)} \bigr\rVert^{2} ,
\qquad
w^{(l)}_i \;=\; 1 + \frac{a^{(l)}_i}{\bar a^{(l)}} ,
\end{equation}
with $a^{(0)} = 0$, $\gamma = \tfrac{1}{2}$ and $\bar a^{(l)} = \tfrac{1}{N}\sum_i a^{(l)}_i$ the mean over tokens. Boundary tokens then pull the next layer's centres toward them, and each layer places its cuts where the previous ones did not (\cref{fig:balanced_tiling}). Note that nothing refers to a particular grid: the input is the coordinate array, so the same construction can apply to 2-D and 3-D grids and point clouds alike. \Cref{sec:exp_local} compares multiple local attention methods and finds that local attention using our tiles with this optimised placement alone already leads to Pareto-optimal results.

\subsection{Faster VQ-Precompute Kernels}
\label{sec:method_pipeline}

Before any query tile runs, one pass over the keys assigns each key to a code and accumulates the per-code aggregates. Its cost is dominated not by the assignment but by the accumulation: every key scatter-adds its $D$-wide value into its code's $\bar V_m$, and since keys anywhere in the sequence can land in the same code, those adds have to be atomic --- each one reads the accumulator, adds to it and writes it back, serialised against every other key landing in that code.

We fix this by first sorting the keys by their assignment. The sort can be an efficient $\mathcal{O}(N)$ counting sort, because we \emph{can} already compute the count aggregates $n_m$ cheaply: they are single integers rather than $D$-dimensional vectors. Ordering the buckets so that a parent's children sit next to each other then puts every code's keys consecutively in that order, and a single pass over the sorted indices can accumulate in registers, writing each output directly once.

This matters more once $\bar K_m$ is needed alongside $\bar V_m$. Each key then carries two $D$-wide payloads into the aggregation instead of one, so an atomic pipeline pays its dominant cost twice. The sorted traversal pays the sorting work once and amortizes it over both $\bar V_m$ and $\bar K_m$, which the same pass accumulates side by side. Against the original atomic kernel, the sorted precompute is about $1.5\times$ faster when only $\bar V_m$ is accumulated and about $2.7\times$ once $\bar K_m$ is accumulated alongside; the precompute is about a third of the layer's time (\cref{app:vq_pipeline}). \Cref{app:kernels} gives both kernels.

\section{Experiments}
\label{sec:experiments}

We evaluate on three datasets: semantic segmentation on ADE20K, classification on ImageNet-1k, and video classification on Diving-48. We evaluate each configuration as a drop-in and after fine-tuning. As a drop-in, the codebook is only fit to the keys of a few batches passed through the pretrained model and no parameter is trained. Fine-tuning uses a training setup that starts from that same codebook initialization and trains on the task loss plus a per-layer distillation term from the original attention, at a budget under $0.05\%$ of the inputs those weights were trained on \citep{ranftl2021dpt,dosovitskiy2021vit,steiner2022augreg,murlabadia2026vjepa21} (\cref{app:training_details}). Every time we report is measured on a single RTX 3090 (\cref{app:benchmark_details}).

We also compare against three efficient-attention methods that do not use vector quantization: MRA attention \citep{zeng2022mra}, Agent Attention \citep{han2024agent} and mixture-of-top-$k$ attention \citep{wen2026mita}. \Cref{tab:baselines} reports one setting of each; the remaining settings are given in \cref{app:baselines}.

\subsection{ADE20K Segmentation and ImageNet-1k Classification}
\label{sec:exp_images}

We replace every attention layer of DPT-Large at $480^2$ ($N = 901$) on ADE20K, and the self-attention of a ViT-B/8 at $224^2$ ($N = 785$) on ImageNet-1k. Both are evaluated as a drop-in and after fine-tuning: ten epochs on ADE20K, one on ImageNet. The pretrained DPT-Large scores 45.88 mean IoU in our pipeline and 46.54 after the same fine-tuning, and we report both; the pretrained ViT-B/8 scores its published 85.82 top-1 and does not benefit from fine-tuning.

\begin{figure}[t]
\begin{center}
\includegraphics[width=\linewidth]{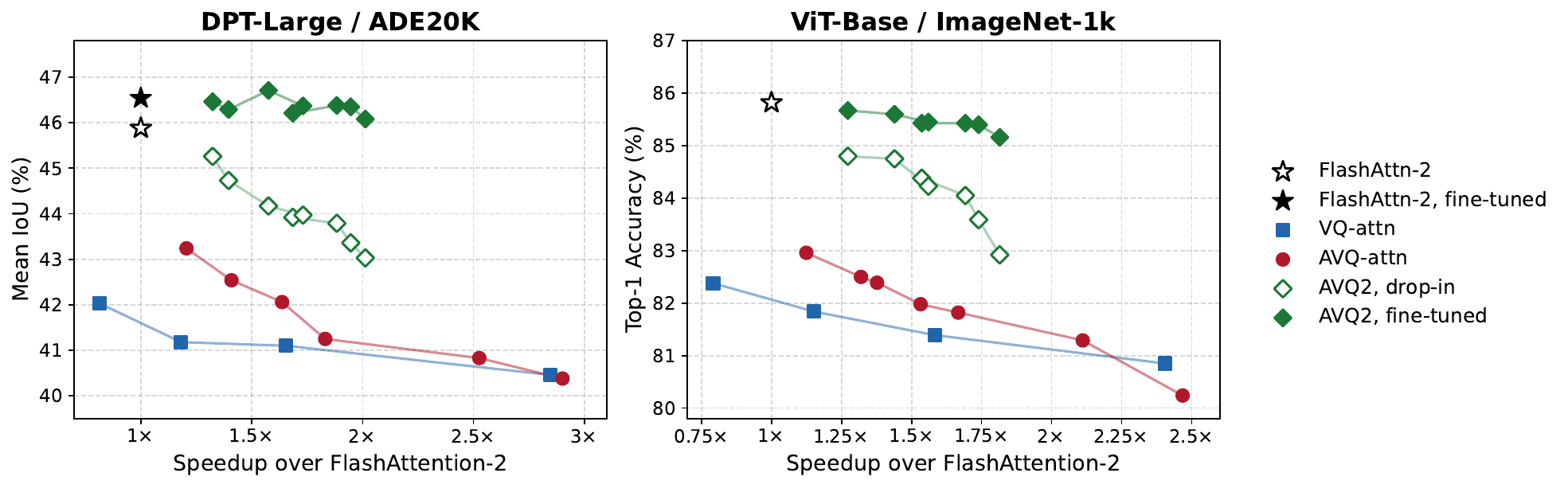}
\end{center}
\caption{AVQ-Attention-2 against other VQ-attention variants
\citep{lingletransformer,vandendool2026avq}: quality against the attention speedup over
FlashAttention-2. Without any fine-tuning
AVQ-Attention-2 lies above both of their fine-tuned curves, so the approximation itself is closer to dense
attention. With fine-tuning it approaches the dense model, at a
$1.3$--$2.0\times$ speedup.}
\label{fig:cost}
\end{figure}

\Cref{fig:cost} reports every configuration we train against its attention speedup over FlashAttention-2, on ADE20K (left) and ImageNet (right). As a drop-in the layer stays within $3$ points of the baseline on both, and the gap closes with codebook size; after fine-tuning it loses little to no quality on ADE20K and comes within $0.7$ points on ImageNet, at a $1.3$--$2.0\times$ and $1.3$--$1.8\times$ speedup respectively. \Cref{tab:baselines} puts three efficient-attention baselines in the same setting: each of these is slower than AVQ-Attention-2 and scores below it on both tasks. Both tasks are small-$N$, where a linear-cost layer has the least to gain; the advantage grows with $N$.

\begin{table}[t]
\caption{Other efficient attention methods, as drop-in and after fine-tuning, under the same recipe. Agent Attention cannot be dropped in: its agent parameters are new and untrained. More details and additional results are in \cref{app:baselines}.}
\label{tab:baselines}
\begin{center}
\small
\setlength{\tabcolsep}{4pt}
\begin{tabular}{lcccccccc}
\toprule
& \multicolumn{4}{c}{ADE20K (mIoU)} & \multicolumn{4}{c}{ImageNet-1k (top-1)} \\
\cmidrule(lr){2-5}\cmidrule(lr){6-9}
& setting & ms & drop-in & fine-tuned & setting & ms & drop-in & fine-tuned \\
\midrule
FlashAttention-2        & ---          & 0.280 & 45.88 & 46.54 & ---        & 0.180 & 85.82 & ---   \\
\midrule
MRA                     & $b=8$        & 1.566 & 31.55 & 45.88 & $b=8$      & 1.041 & 83.64 & 85.32 \\
Agent Attention         & $15\times15$ & 0.576 & ---   & 44.64 & $7\times7$ & 0.183 & ---   & 77.27 \\
Mixture-of-top-$k$      & $7\times7$   & 2.162 & 21.38 & 46.17 & $7\times7$ & 2.139 & 79.95 & 84.76 \\
\midrule
AVQ-Attention-2         & $32/8/8$     & \textbf{0.178} & \textbf{44.17} & \textbf{46.71} & $32/8/8$   & \textbf{0.125} & \textbf{84.75} & \textbf{85.60} \\
\bottomrule
\end{tabular}
\end{center}
\end{table}

\textbf{Building up from AVQ-attention} separates our changes from previous work: \cref{tab:ablation} adds one component at a time. Cluster means and exact local attention each help on their own and are best together. The difference is sharpest without any fine-tuning at all: with the codebook fitted and no gradient step taken, AVQ-attention collapses where AVQ2 loses only a few points.

\begin{table}[t]
\caption{Building the layer up from AVQ-attention. ADE20K mean IoU (DPT-Large, 10 epochs) and ImageNet-1k top-1 (ViT-B/8, 1 epoch) after fine-tuning, at two codebook configurations $M_0/\mathcal{P}/\mathcal{C}$; the published row is at \citeauthor{vandendool2026avq}'s budget (4 instead of our 1 epoch on ImageNet); every other row is ours, cumulative from that row down. Without any fine-tuning AVQ2 already reaches most of its accuracy, where AVQ collapses entirely.}
\label{tab:ablation}
\begin{center}
\begin{tabular}{lcccc}
\toprule
& \multicolumn{2}{c}{ADE20K (mIoU)} & \multicolumn{2}{c}{ImageNet-1k (top-1)} \\
\cmidrule(lr){2-3}\cmidrule(lr){4-5}
& $32/8/4$ & $32/8/8$ & $32/8/4$ & $32/8/8$ \\
\midrule
FlashAttention-2 baseline               & 45.88 & 45.88 & 85.82 & 85.82 \\
FlashAttention-2, fine-tuned            & 46.54 & 46.54 & ---   & ---   \\
\midrule
AVQ-attention, as published & 40.38 & 40.83 & 80.24 & 81.29 \\
AVQ, our training setup                 & 41.74 & 43.09 & 73.35 & 76.58 \\
\quad $+$ cluster means                   & 43.36 & 44.47 & 80.74 & 82.43 \\
\quad $+$ exact local attention         & 45.49 & 45.35 & 84.56 & 84.77 \\
\quad $+$ both $=$ AVQ2                 & \textbf{46.38} & \textbf{46.71} & \textbf{85.43} & \textbf{85.60} \\
\midrule
\multicolumn{5}{l}{\textit{Drop-in, no training}} \\
AVQ                         & 13.98 & 18.22 & 5.59  & 12.55 \\
AVQ2                        & 43.79 & 44.17 & 84.05 & 84.75 \\
\bottomrule
\end{tabular}
\end{center}
\end{table}

\begin{figure}[t]
\begin{center}
\includegraphics[width=\linewidth]{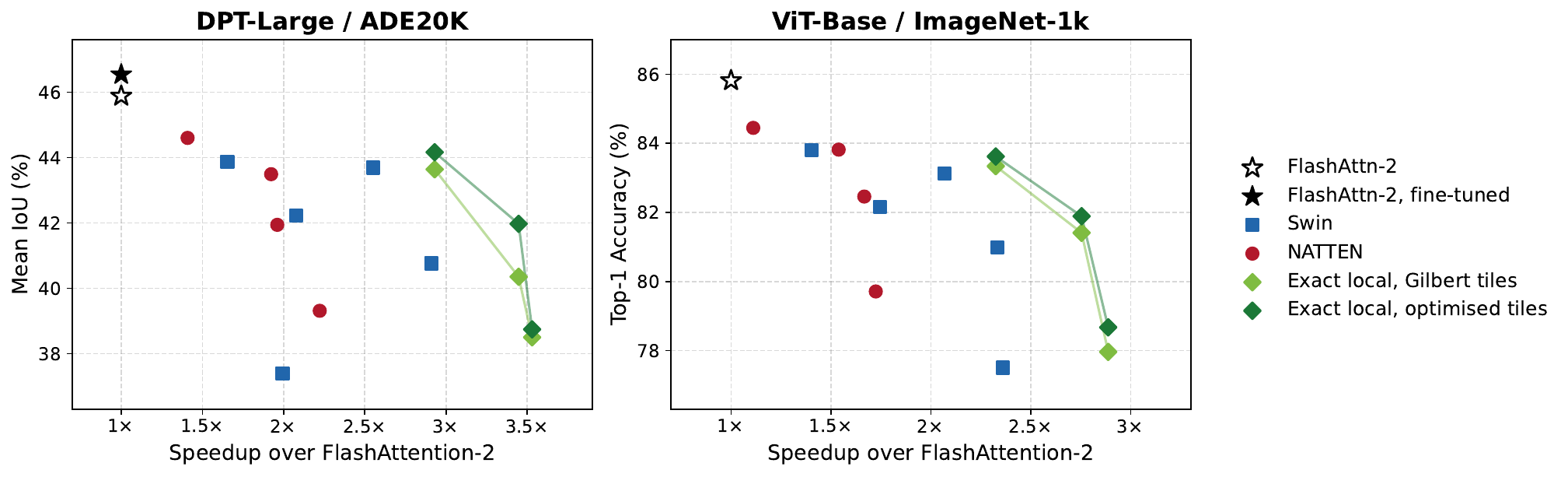}
\end{center}
\caption{Local attention alone: quality after fine-tuning under the setup of
\cref{app:training_details}, against the attention speedup over FlashAttention-2.
Optimised tiles beat Gilbert tiles at every setting, showing how the seemingly simple decision of how to
divide tokens over tiles already gives a free win. Only our tilings are drawn as curves: Swin's speed is
not monotone in its window size, which has to do with grid padding and kernel artifacts, so its points do
not order along the axis. The settings behind every point, and the exact numbers, are in \cref{tab:local}.}
\label{fig:local}
\end{figure}

\label{sec:exp_local}\textbf{Exact local attention}, with no AVQ-attention involved, is compared against other local attention methods in \cref{fig:local}: our tiles, in the Gilbert tiling of \citet{vandendool2026avq} and the optimised tiling of \cref{sec:method}, against Swin \citep{liu2021swin} and NATTEN \citep{hassani2023nat} under the same training setup. Our exact local attention based on optimised tiles is Pareto-optimal against the other baselines on both tasks.

\textbf{Other ablations} are in \cref{app:refine_criterion,app:refine_ablation}, on how a tile selects the parents it refines, and in \cref{app:count_correction}, on skipping the correction of \cref{sec:method_localexact} that keeps a tile's local keys from counting twice.

\subsection{Video at Larger Sequence Lengths}
\label{sec:exp_video}

We scale to video with V-JEPA~2 \citep{assran2025vjepa2}, where different frames per clip give $N = 9216$ and $18432$ from one set of weights. We replace self-attention in every block of the ViT-B/16 encoder and follow V-JEPA~2's protocol: the encoder is frozen and an \emph{attentive probe} --- a small attention-pooling classifier over its output tokens --- is trained on top to determine the encoder's quality. The replaced layers are fine-tuned by distillation on unlabelled clips.

\Cref{tab:video} reports classification accuracy and the cost per attention layer; over the whole encoder at this length the speedup is $3.5\times$. Dropped in, with no training at all, our layer is level with dense attention at the longest setting; after distillation it stays within a point of dense. Furthermore we use this experiment setting to verify that AVQ-Attention-2 can train from scratch: replacing the attention of the probe instead of that of the encoder, and training it from initialization, lands in the same range.

\begin{figure}[t]
\begin{center}
\includegraphics[width=\linewidth]{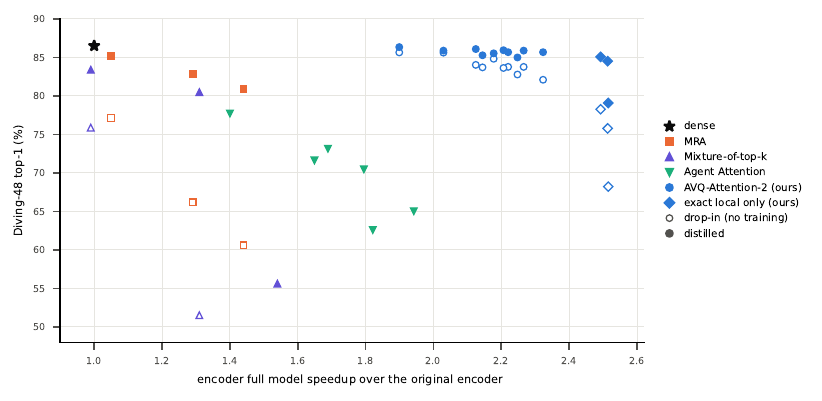}
\end{center}
\caption{Diving-48 at $N = 9216$: top-1 probe accuracy of the V-JEPA~2 encoder against its throughput
speedup over the original encoder. \Cref{app:vjepa_configs} shows our own configurations zoomed in, as
well as the hyperparameters used for each point.}
\label{fig:vjepa9216}
\end{figure}

\begin{table}[t]
\caption{Diving-48 at $N = 18432$, a V-JEPA~2 ViT-B/16 encoder with an attentive probe. On the left,
every attention layer of the encoder is replaced, as a drop-in and after distillation, beside the cost of
one encoder attention layer. On the right, the encoder is left dense and the probe is trained from scratch with an efficient attention method, beside the cost of one probe layer forward and backward. Every setting is in \cref{app:baselines}.}
\label{tab:video}
\begin{center}
\small
\begin{minipage}[t]{0.57\linewidth}
\centering
\begin{tabular}{lccc}
\toprule
\multicolumn{4}{c}{\textit{Encoder}} \\
\cmidrule(lr){1-4}
& drop-in & distilled & ms \\
\midrule
Dense              & 88.83 & --- & 15.9 \\
AVQ-Attention-2    & 88.73 & 88.22 & 0.94 \\
Mixture-of-top-$k$ & 73.10 & 85.58 & 5.85 \\
Agent Attention    & ---   & 82.54 & 4.02 \\
MRA                & 79.90 & 86.90 & 7.53 \\
\bottomrule
\end{tabular}
\end{minipage}
\hfill
\begin{minipage}[t]{0.40\linewidth}
\centering
\begin{tabular}{lcc}
\toprule
\multicolumn{3}{c}{\textit{Probe, trained from scratch}} \\
\cmidrule(lr){1-3}
& top-1 & fwd\,+\,bwd ms \\
\midrule
Dense              & 88.83 & 74.3 \\
AVQ-Attention-2    & 90.05 & 11.1 \\
Mixture-of-top-$k$ & 89.49 & 16.5 \\
Agent Attention    & 89.44 & 46.2 \\
MRA                & 86.75 & 15.1 \\
\bottomrule
\end{tabular}
\end{minipage}
\end{center}
\end{table}

\section{Discussion and Conclusion}
\label{sec:conclusion}

We have presented AVQ-Attention-2, improving on AVQ-attention in three ways: faster training, the possibility to drop it into existing networks with little to no fine-tuning, and Pareto-optimal results, competitive with FlashAttention already at sequence lengths as short as $N=785$.

The experiments test the layer as a general-purpose replacement rather than a benchmark-specific one: we take pretrained transformers, replace every attention layer, and fine-tune briefly or not at all. The margin grows with sequence length. In a video encoder at $18432$ tokens both a forward layer and a training step of it cost a fraction of what FlashAttention costs, at an accuracy within a point of dense attention. Exact attention grows quadratically in $N$ where ours grows linearly, so that gap widens for the larger $N$ settings.

As future work we plan to move to higher resolutions, longer video generation, and the decoder setting, where a causal variant of the layer would meet much longer sequences. We showcase AVQ-Attention-2 as an efficient attention layer only, but at larger sequence lengths attention makes up a larger share of the model, and the claim can move from the layer to the throughput of the whole model, aiming for results on specific tasks rather than for a general-purpose replacement.

\section*{Acknowledgements}
This work is financially supported by Qualcomm Technologies Inc., the University of Amsterdam, and the allowance Top consortia for Knowledge and Innovation from the Netherlands Ministry of Economic Affairs and Climate Policy.

\bibliography{main}

@String(CVPR  = {IEEE Conf. Comput. Vis. Pattern Recog.})

@String(ICCV  = {Int. Conf. Comput. Vis.})

@String(ECCV  = {Eur. Conf. Comput. Vis.})

@String(NeurIPS = {Adv. Neural Inform. Process. Syst.})

@String(ICML  = {Int. Conf. Mach. Learn.})

@String(ICLR  = {Int. Conf. Learn. Represent.})

@String(BMVC  = {Brit. Mach. Vis. Conf.})

@String(CVPR  = {CVPR})

@String(ICCV  = {ICCV})

@String(ECCV  = {ECCV})

@String(NeurIPS = {NeurIPS})

@String(ICML  = {ICML})

@String(ICLR  = {ICLR})

@String(BMVC  =	{BMVC})

@inproceedings{dao2022flashattention,
  title     = {{FlashAttention}: Fast and Memory-Efficient Exact Attention with {IO}-Awareness},
  author    = {Dao, Tri and Fu, Daniel Y. and Ermon, Stefano and Rudra, Atri and R{\'e}, Christopher},
  booktitle = {Advances in Neural Information Processing Systems},
  volume    = {35},
  year      = {2022},
  url       = {https://arxiv.org/abs/2205.14135}
}

@article{beltagy2020longformerlongdocumenttransformer,
  title   = {Longformer: The Long-Document Transformer},
  author  = {Beltagy, Iz and Peters, Matthew E. and Cohan, Arman},
  journal = {arXiv preprint},
  volume  = {2004.05150},
  year    = {2020},
  url     = {https://arxiv.org/abs/2004.05150}
}

@article{child2019generatinglongsequencessparse,
  title   = {Generating Long Sequences with Sparse Transformers},
  author  = {Child, Rewon and Gray, Scott and Radford, Alec and Sutskever, Ilya},
  journal = {arXiv preprint},
  volume  = {1904.10509},
  year    = {2019},
  url     = {https://arxiv.org/abs/1904.10509}
}

@article{roy2021efficient,
  title={Efficient content-based sparse attention with routing transformers},
  author={Roy, Aurko and Saffar, Mohammad and Vaswani, Ashish and Grangier, David},
  journal={Transactions of the Association for Computational Linguistics},
  volume={9},
  pages={53--68},
  year={2021}
}

@inproceedings{lingletransformer,
  title={{Transformer-VQ}: Linear-Time Transformers via Vector Quantization},
  author={Lingle, Lucas D},
  booktitle={International Conference on Learning Representations},
  year={2024}
}

@inproceedings{yuan2025native,
  title={Native sparse attention: Hardware-aligned and natively trainable sparse attention},
  author={Yuan, Jingyang and Gao, Huazuo and Dai, Damai and Luo, Junyu and Zhao, Liang and Zhang, Zhengyan and Xie, Zhenda and Wei, Yuxing and Wang, Lean and Xiao, Zhiping and others},
  booktitle={Proceedings of the 63rd Annual Meeting of the Association for Computational Linguistics (Volume 1: Long Papers)},
  pages={23078--23097},
  year={2025}
}

@inproceedings{bolya2022tokenmerging,
  title        = {Token Merging: Your {ViT} But Faster},
  author       = {Bolya, Daniel and Fu, Cheng-Yang and Dai, Xiaoliang and Zhang, Peizhao and Feichtenhofer, Christoph and Hoffman, Judy},
  booktitle    = {International Conference on Learning Representations (ICLR)},
  year         = {2023},
  note         = {Oral presentation},
  url          = {https://arxiv.org/abs/2210.09461},
}

@inproceedings{dosovitskiy2021vit,
  title     = {An Image is Worth 16x16 Words: Transformers for Image Recognition at Scale},
  author    = {Dosovitskiy, Alexey and Beyer, Lucas and Kolesnikov, Alexander and Weissenborn, Dirk and Zhai, Xiaohua and Unterthiner, Thomas and Dehghani, Mostafa and Minderer, Matthias and Heigold, Georg and Gelly, Sylvain and Uszkoreit, Jakob and Houlsby, Neil},
  booktitle = ICLR,
  year      = {2021},
  url       = {https://arxiv.org/abs/2010.11929}
}

@inproceedings{dao2024flashattention2,
  title     = {{FlashAttention-2}: Faster Attention with Better Parallelism and Work Partitioning},
  author    = {Dao, Tri},
  booktitle = ICLR,
  year      = {2024},
  url       = {https://arxiv.org/abs/2307.08691}
}

@inproceedings{tillet2019triton,
  title     = {Triton: An Intermediate Language and Compiler for Tiled Neural Network Computations},
  author    = {Tillet, Philippe and Kung, H. T. and Cox, David},
  booktitle = {Proceedings of the 3rd ACM SIGPLAN International Workshop on Machine Learning and Programming Languages (MAPL)},
  pages     = {10--19},
  year      = {2019}
}

@inproceedings{katharopoulos2020linear,
  title     = {Transformers are {RNNs}: Fast Autoregressive Transformers with Linear Attention},
  author    = {Katharopoulos, Angelos and Vyas, Apoorv and Pappas, Nikolaos and Fleuret, Fran{\c{c}}ois},
  booktitle = ICML,
  year      = {2020},
  url       = {https://arxiv.org/abs/2006.16236}
}

@inproceedings{deng2009imagenet,
  title     = {{ImageNet}: A Large-Scale Hierarchical Image Database},
  author    = {Deng, Jia and Dong, Wei and Socher, Richard and Li, Li-Jia and Li, Kai and Fei-Fei, Li},
  booktitle = CVPR,
  pages     = {248--255},
  year      = {2009}
}

@article{wang2020linformer,
  title   = {Linformer: Self-Attention with Linear Complexity},
  author  = {Wang, Sinong and Li, Belinda Z. and Khabsa, Madian and Fang, Han and Ma, Hao},
  journal = {arXiv preprint arXiv:2006.04768},
  year    = {2020},
  url     = {https://arxiv.org/abs/2006.04768}
}

@misc{cerveny2018gilbert,
  title        = {Gilbert: Generalized {Hilbert} (``{Gilbert}'') space-filling curve for rectangular domains of arbitrary (non-power of two) sizes},
  author       = {\v{C}erven\'{y}, Jakub},
  year         = {2018},
  howpublished = {\url{https://github.com/jakubcerveny/gilbert}},
  note         = {GitHub repository}
}

@inproceedings{ranftl2021dpt,
  title     = {Vision Transformers for Dense Prediction},
  author    = {Ranftl, Ren{\'e} and Bochkovskiy, Alexey and Koltun, Vladlen},
  booktitle = ICCV,
  pages     = {12179--12188},
  year      = {2021},
  url       = {https://arxiv.org/abs/2103.13413}
}

@inproceedings{zhou2017ade20k,
  title     = {Scene Parsing through {ADE20K} Dataset},
  author    = {Zhou, Bolei and Zhao, Hang and Puig, Xavier and Fidler, Sanja and Barriuso, Adela and Torralba, Antonio},
  booktitle = CVPR,
  pages     = {633--641},
  year      = {2017}
}

@inproceedings{hassani2023nat,
  title     = {Neighborhood Attention Transformer},
  author    = {Hassani, Ali and Walton, Steven and Li, Jiachen and Li, Shen and Shi, Humphrey},
  booktitle = CVPR,
  year      = {2023},
  url       = {https://arxiv.org/abs/2204.07143}
}

@inproceedings{liu2021swin,
  title     = {{Swin Transformer}: Hierarchical Vision Transformer using Shifted Windows},
  author    = {Liu, Ze and Lin, Yutong and Cao, Yue and Hu, Han and Wei, Yixuan and Zhang, Zheng and Lin, Stephen and Guo, Baining},
  booktitle = ICCV,
  year      = {2021},
  url       = {https://arxiv.org/abs/2103.14030}
}

@inproceedings{choromanski2021performer,
  title     = {Rethinking Attention with {Performers}},
  author    = {Choromanski, Krzysztof and Likhosherstov, Valerii and Dohan, David and Song, Xingyou and Gane, Andreea and Sarl{\'o}s, Tam{\'a}s and Hawkins, Peter and Davis, Jared and Mohiuddin, Afroz and Kaiser, {\L}ukasz and Belanger, David and Colwell, Lucy and Weller, Adrian},
  booktitle = ICLR,
  year      = {2021},
  url       = {https://arxiv.org/abs/2009.14794}
}

@article{alonso2026ovq,
  title   = {Online Vector Quantized Attention},
  author  = {Alonso, Nick and Figliolia, Tomas and Millidge, Beren},
  journal = {arXiv preprint},
  volume  = {2602.03922},
  year    = {2026},
  url     = {https://arxiv.org/abs/2602.03922}
}

@inproceedings{kong2023goat,
  title     = {{GOAT}: A Global Transformer on Large-scale Graphs},
  author    = {Kong, Kezhi and Chen, Jiuhai and Kirchenbauer, John and Ni, Renkun and Bruss, C. Bayan and Goldstein, Tom},
  booktitle = {Proc. International Conference on Machine Learning (ICML)},
  series    = {Proceedings of Machine Learning Research},
  volume    = {202},
  pages     = {17375--17390},
  year      = {2023},
  publisher = {PMLR},
  url       = {https://proceedings.mlr.press/v202/kong23a.html}
}

@inproceedings{zhou2026svgear,
  title     = {{SVG-EAR}: Parameter-Free Linear Compensation for Sparse Video Generation via Error-aware Routing},
  author    = {Zhou, Xuanyi and Mang, Qiuyang and Yang, Shuo and Xi, Haocheng and Zhang, Jintao and Mao, Huanzhi and Gonzalez, Joseph E. and Keutzer, Kurt and Stoica, Ion and Cheung, Alvin},
  booktitle = {European Conference on Computer Vision (ECCV)},
  pages     = {167--185},
  year      = {2026},
  publisher = {Springer},
  doi       = {10.1007/978-3-032-37359-5_10},
  url       = {https://arxiv.org/abs/2603.08982}
}

@inproceedings{li2025commvq,
  title     = {{CommVQ}: Commutative Vector Quantization for {KV} Cache Compression},
  author    = {Li, Junyan and Zhang, Yang and Hassan, Muhammad Yusuf and Chafekar, Talha and Cai, Tianle and Ren, Zhile and Guo, Pengsheng and Karimzadeh, Foroozan and Reed, Colorado and Wang, Chong and Gan, Chuang},
  booktitle = {Proc. International Conference on Machine Learning (ICML)},
  year      = {2025},
  url       = {https://arxiv.org/abs/2506.18879}
}

@inproceedings{hooper2024kvquant,
  title     = {{KVQuant}: Towards 10 Million Context Length {LLM} Inference with {KV} Cache Quantization},
  author    = {Hooper, Coleman and Kim, Sehoon and Mohammadzadeh, Hiva and Mahoney, Michael W. and Shao, Yakun Sophia and Keutzer, Kurt and Gholami, Amir},
  booktitle = {Advances in Neural Information Processing Systems (NeurIPS)},
  year      = {2024},
  url       = {https://arxiv.org/abs/2401.18079}
}

@inproceedings{vandendool2026avq,
  title     = {{AVQ-Attention}: {Adaptive} {Vector-Quantized} {Attention}},
  author    = {van den Dool, Winfried and Forr{\'e}, Patrick and Habibian, Amir and Asano, Yuki M. and Welling, Max},
  booktitle = {European Conference on Computer Vision (ECCV)},
  year      = {2026},
  url       = {https://arxiv.org/abs/2607.12789}
}

@inproceedings{gu2022s4,
  title     = {Efficiently Modeling Long Sequences with Structured State Spaces},
  author    = {Gu, Albert and Goel, Karan and R{\'e}, Christopher},
  booktitle = {International Conference on Learning Representations (ICLR)},
  year      = {2022},
  url       = {https://arxiv.org/abs/2111.00396}
}

@inproceedings{gu2024mamba,
  title     = {Mamba: Linear-Time Sequence Modeling with Selective State Spaces},
  author    = {Gu, Albert and Dao, Tri},
  booktitle = {Conference on Language Modeling (COLM)},
  year      = {2024},
  url       = {https://arxiv.org/abs/2312.00752}
}

@article{assran2025vjepa2,
  title   = {{V-JEPA} 2: Self-Supervised Video Models Enable Understanding, Prediction and Planning},
  author  = {Assran, Mido and Bardes, Adrien and Fan, David and Garrido, Quentin and Howes, Russell and Komeili, Mojtaba and Muckley, Matthew and Rizvi, Ammar and Roberts, Claire and Sinha, Koustuv and Zholus, Artem and Arnaud, Sergio and Gejji, Abha and Martin, Ada and Hogan, Francois Robert and Dugas, Daniel and Bojanowski, Piotr and Khalidov, Vasil and Labatut, Patrick and Massa, Francisco and Szafraniec, Marc and Krishnakumar, Kapil and Li, Yong and Ma, Xiaodong and Chandar, Sarath and Meier, Franziska and LeCun, Yann and Rabbat, Michael and Ballas, Nicolas},
  journal = {arXiv preprint},
  volume  = {2506.09985},
  year    = {2025},
  url     = {https://arxiv.org/abs/2506.09985}
}

@inproceedings{li2018diving48,
  title     = {{RESOUND}: Towards Action Recognition without Representation Bias},
  author    = {Li, Yingwei and Li, Yi and Vasconcelos, Nuno},
  booktitle = {Proc. European Conference on Computer Vision (ECCV)},
  year      = {2018},
  url       = {https://openaccess.thecvf.com/content_ECCV_2018/html/Yingwei_Li_RESOUND_Towards_Action_ECCV_2018_paper.html}
}

@techreport{bradley2000constrained,
  title       = {Constrained K-Means Clustering},
  author      = {Bradley, Paul S. and Bennett, Kristin P. and Demiriz, Ayhan},
  institution = {Microsoft Research},
  number      = {MSR-TR-2000-65},
  year        = {2000},
}

@inproceedings{yu2022boat,
  title     = {{BOAT}: {Bilateral} {Local} {Attention} {Vision} {Transformer}},
  author    = {Yu, Tan and Zhao, Gangming and Li, Ping and Yu, Yizhou},
  booktitle = {British Machine Vision Conference (BMVC)},
  year      = {2022},
  url       = {https://arxiv.org/abs/2201.13027}
}

@inproceedings{zhang2022vsa,
  title     = {{VSA}: {Learning} {Varied-Size} {Window} {Attention} in {Vision} {Transformers}},
  author    = {Zhang, Qiming and Xu, Yufei and Zhang, Jing and Tao, Dacheng},
  booktitle = {European Conference on Computer Vision (ECCV)},
  year      = {2022},
  url       = {https://arxiv.org/abs/2204.08446}
}

@inproceedings{weissenborn2020scaling,
  title     = {Scaling Autoregressive Video Models},
  author    = {Weissenborn, Dirk and T{\"a}ckstr{\"o}m, Oscar and Uszkoreit, Jakob},
  booktitle = {International Conference on Learning Representations (ICLR)},
  year      = {2020},
  url       = {https://arxiv.org/abs/1906.02634}
}

@inproceedings{zhang2025sage,
  title     = {{SageAttention}: Accurate 8-Bit Attention for Plug-and-play Inference Acceleration},
  author    = {Zhang, Jintao and Wei, Jia and Huang, Haofeng and Zhang, Pengle and Zhu, Jun and Chen, Jianfei},
  booktitle = {International Conference on Learning Representations (ICLR)},
  year      = {2025},
  url       = {https://arxiv.org/abs/2410.02367}
}

@article{kang2025fma,
  title     = {Fast Multipole Attention: A Scalable Multilevel Attention Mechanism for Text and Images},
  author    = {Kang, Yanming and Tran, Giang and De Sterck, Hans},
  journal   = {Neural Networks},
  volume    = {206},
  pages     = {109847},
  year      = {2026},
  doi       = {10.1016/j.neunet.2026.109847},
  url       = {https://arxiv.org/abs/2310.11960}
}

@inproceedings{zhu2021htransformer,
  title     = {{H-Transformer-1D}: Fast One-Dimensional Hierarchical Attention for Sequences},
  author    = {Zhu, Zhenhai and Soricut, Radu},
  booktitle = {Proceedings of the 59th Annual Meeting of the Association for Computational Linguistics and the 11th International Joint Conference on Natural Language Processing (Volume 1: Long Papers)},
  pages     = {3801--3815},
  year      = {2021},
  publisher = {Association for Computational Linguistics},
  doi       = {10.18653/v1/2021.acl-long.294},
  url       = {https://aclanthology.org/2021.acl-long.294/}
}

@inproceedings{zeng2022mra,
  title     = {Multi Resolution Analysis ({MRA}) for Approximate Self-Attention},
  author    = {Zeng, Zhanpeng and Pal, Sourav and Kline, Jeffery and Fung, Glenn M. and Singh, Vikas},
  booktitle = {International Conference on Machine Learning (ICML)},
  series    = {Proceedings of Machine Learning Research},
  volume    = {162},
  pages     = {25955--25972},
  year      = {2022},
  publisher = {PMLR},
  url       = {https://proceedings.mlr.press/v162/zeng22a.html}
}

@inproceedings{han2024agent,
  title     = {Agent Attention: On the Integration of Softmax and Linear Attention},
  author    = {Han, Dongchen and Ye, Tianzhu and Han, Yizeng and Xia, Zhuofan and Pan, Siyuan and Wan, Pengfei and Song, Shiji and Huang, Gao},
  booktitle = {European Conference on Computer Vision (ECCV)},
  year      = {2024},
  url       = {https://arxiv.org/abs/2312.08874}
}

@inproceedings{wen2026mita,
  title     = {Mixture-of-{T}op-$k$ Attention: Efficient Attention via Scalable Fast Weights},
  author    = {Wen, Qishuai and Huang, Zhiyuan and Meng, Xianghan and He, Wei and Li, Chun-Guang},
  booktitle = {Advances in Neural Information Processing Systems (NeurIPS)},
  year      = {2026},
  eprint    = {2602.01219},
  archivePrefix = {arXiv},
  primaryClass  = {cs.LG},
  url       = {https://arxiv.org/abs/2602.01219}
}

@inproceedings{chen2021scatterbrain,
  title     = {Scatterbrain: Unifying Sparse and Low-rank Attention},
  author    = {Chen, Beidi and Dao, Tri and Winsor, Eric and Song, Zhao and Rudra, Atri and R\'{e}, Christopher},
  booktitle = {Advances in Neural Information Processing Systems (NeurIPS)},
  volume    = {34},
  pages     = {17413--17426},
  year      = {2021},
  url       = {https://proceedings.neurips.cc/paper_files/paper/2021/hash/9185f3ec501c674c7c788464a36e7fb3-Abstract.html}
}

@inproceedings{lu2025moba,
  title     = {{MoBA}: Mixture of Block Attention for Long-Context {LLMs}},
  author    = {Lu, Enzhe and Jiang, Zhejun and Liu, Jingyuan and Du, Yulun and Jiang, Tao and Hong, Chao and Liu, Shaowei and He, Weiran and Yuan, Enming and Wang, Yuzhi and Huang, Zhiqi and Yuan, Huan and Xu, Suting and Xu, Xinran and Lai, Guokun and Chen, Yanru and Zheng, Huabin and Yan, Junjie and Su, Jianlin and Wu, Yuxin and Zhang, Yutao and Yang, Zhilin and Zhou, Xinyu and Zhang, Mingxing and Qiu, Jiezhong},
  booktitle = {Advances in Neural Information Processing Systems (NeurIPS)},
  volume    = {38},
  pages     = {20267--20292},
  year      = {2025},
  url       = {https://arxiv.org/abs/2502.13189}
}

@article{steiner2022augreg,
  title   = {How to Train Your {ViT}? Data, Augmentation, and Regularization in Vision Transformers},
  author  = {Steiner, Andreas and Kolesnikov, Alexander and Zhai, Xiaohua and Wightman, Ross and Uszkoreit, Jakob and Beyer, Lucas},
  journal = {Transactions on Machine Learning Research},
  year    = {2022},
  url     = {https://arxiv.org/abs/2106.10270}
}

@article{murlabadia2026vjepa21,
  title   = {{V-JEPA} 2.1: Unlocking Dense Features in Video Self-Supervised Learning},
  author  = {Mur-Labadia, Lorenzo and Muckley, Matthew and Bar, Amir and Assran, Mido and Sinha, Koustuv and Rabbat, Mike and LeCun, Yann and Ballas, Nicolas and Bardes, Adrien},
  journal = {arXiv preprint},
  volume  = {2603.14482},
  year    = {2026},
  url     = {https://arxiv.org/abs/2603.14482}
}
\bibliographystyle{iclr2027_conference}

\newpage
\appendix
\section{Training Details}
\label{app:training_details}

\paragraph{Segmentation (ADE20K).}
DPT-Large (\texttt{Intel/dpt-large-ade}) on \texttt{scene\_parse\_150}, every attention layer replaced. Images are $480^2$, giving $N = 901$ tokens. We train 10 epochs with an effective batch size of 16, AdamW at learning rate $10^{-5}$, linear warmup over the first two epochs and a constant rate thereafter, and per-layer distillation. Our evaluation pipeline is not identical to the one the released checkpoint was produced under, so we also fine-tune the unmodified model with this same setup and report both references. Training augmentation resizes to $480s$ for $s \in \{1, 1.25, 1.5, 1.75, 2\}$, then random-crops to $480$ and flips horizontally; validation resizes to $480^2$. Every fine-tuned entry of \cref{fig:cost} and \cref{tab:ablation} uses this setup.

\paragraph{Classification (ImageNet-1k).}
ViT-B/8 (\texttt{vit\_base\_patch8\_224.augreg\_in21k\_ft\_in1k}) at $224^2$ ($N = 785$), one epoch, effective batch size 256, learning rate warmed up linearly to $5\times10^{-6}$ over the epoch, and the same per-layer distillation. The final block computes only the class-token query, exactly.

\paragraph{Video (Diving-48).}
V-JEPA~2.1 ViT-B/16 at a 384 crop. Tubelets span two frames, so $F$ frames at crop $c$ give $(F/2)\times(c/16)^2$ tokens. Diving-48 \citep{li2018diving48} separates its 48 classes by temporal structure rather than appearance --- every clip shows the same pool and board --- so a dive is recognised only by relating tokens across the whole clip. We use its train and test split as distributed with the dataset. The attentive probe is trained for 20 epochs at an effective batch size of 16, AdamW at learning rate $10^{-3}$ with a cosine schedule and no warmup, weight decay 0.8, and bf16. Adaptation runs for 3 epochs at learning rate $10^{-4}$ without weight decay on unlabelled training clips. We report the peak top-1 over the 20 probe epochs. This protocol is the same at both sequence lengths and for every configuration shown; the configurations themselves differ. \Cref{tab:video} uses $M_0 = 64$, $\mathcal{P} = 16$ and $\mathcal{C} = 8$ with exact local attention over $128$-token tiles, while \cref{fig:vjepa9216} varies the codebook and the tile size over the points named in \cref{app:vjepa_configs}.

\paragraph{Codebook initialization.}
The codebook is initialized from real keys collected by passing a few batches through the pretrained model, at a fixed budget of keys per codeword, so that the amount of initialization data scales with codebook size. A drop-in evaluation runs this step and nothing else: no weight of the model is changed. During fine-tuning the codebook receives no gradient and is maintained by EMA alone, and a cluster that stops receiving keys is revived by splitting the highest-distortion live cluster.

\paragraph{Class token under local attention.}
A class token has no position on the grid, so it belongs to no tile, window or neighbourhood. In every local-attention layer --- our exact-local-only layer, Swin and NATTEN --- it is therefore handled by one extra exact pass per layer: the class-token query attends to all $N$ keys, and the patch queries attend within their tile with the class token's key excluded. The pass is shared by the three implementations and is included in their reported attention time (\cref{app:benchmark_details}). AVQ-attention needs no such pass, since every query, the class token included, reaches the whole sequence through the codebook.

\paragraph{Per-layer distillation.}
Every replaced layer is regressed onto the output of the layer it replaced in the frozen pre-replacement model. The loss is scale-free --- each layer's error is divided by the teacher's own mean square --- and weighted by the teacher's measured task loss so that one weight means the same thing across tasks (1.9411 on ADE20K, 0.5432 on ImageNet). Every fine-tuned result uses it except the AVQ-attention reference row, which is trained as in the original paper; \cref{tab:ablation} separate the two.

\section{Benchmark Details}
\label{app:benchmark_details}
All attention times are measured on a single NVIDIA RTX 3090, in fp16 on ADE20K and ImageNet-1k and in bf16 on Diving-48, and cover both kernels of the layer. The two are captured in one CUDA graph: three eager calls populate the allocator and twenty replays settle the caches, after which two hundred timed replays are recorded, with the events taken outside the graph. Each reported number is the median of five such measurements. Query tiles are $64$ tokens on ADE20K and ImageNet, and $128$ on Diving-48, unless a configuration names another size.

\subsection{Sorted versus atomic aggregation}
\label{app:vq_pipeline}
\Cref{sec:method_pipeline} replaces the per-key atomic scatter of the original AVQ precompute with a counting sort followed by one sorted traversal. \Cref{tab:vq_pipeline} times the two pipelines on the same clustered keys and the same codebook, at one geometry across sequence lengths, accumulating value sums alone and value and key sums together. The atomic kernel is the original one, run at its fastest documented setting. The sorted pipeline is about $1.5\times$ faster on value sums alone and about $2.7\times$ once key sums are accumulated alongside, and the ratio does not depend on the sequence length: adding the second aggregate costs the sorted pass $30\%$ and the atomic one $150\%$, since each atomic scatter is paid once per key per aggregate while the sorted traversal reads each key once and writes each cluster once. At this geometry the precompute is $41\%$ of the full layer at 1k tokens and $32$--$35\%$ from 4k up.

\begin{table}[h]
\caption{Quantization precompute, sorted pipeline against the original atomic scatter, accumulating value sums alone and value and key sums together. Milliseconds on one RTX 3090 in fp16; $B{=}4$, $H{=}16$, $D{=}64$, $M_0{=}64$, $\mathcal{C}{=}8$, depth 1.}
\label{tab:vq_pipeline}
\begin{center}
\begin{tabular}{rcccccc}
\toprule
& \multicolumn{3}{c}{$\bar V_m$ only} & \multicolumn{3}{c}{$\bar V_m$ and $\bar K_m$} \\
\cmidrule(lr){2-4}\cmidrule(lr){5-7}
$N$ & sorted & atomic & ratio & sorted & atomic & ratio \\
\midrule
1{,}024  & 0.079 & 0.112 & $1.4\times$ & 0.106 & 0.276  & $2.6\times$ \\
4{,}096  & 0.240 & 0.400 & $1.7\times$ & 0.304 & 0.827  & $2.7\times$ \\
16{,}384 & 0.936 & 1.543 & $1.6\times$ & 1.113 & 3.043  & $2.7\times$ \\
65{,}536 & 3.697 & 6.172 & $1.7\times$ & 4.326 & 11.985 & $2.8\times$ \\
\bottomrule
\end{tabular}
\end{center}
\end{table}

\section{Efficient-Attention Baselines}
\label{app:baselines}
Every baseline runs in its authors' own implementation, is trained under the recipe of \cref{app:training_details} and is timed as in \cref{app:benchmark_details}. The setting named in \cref{tab:baselines} is the block size $b$ for MRA, the agent grid for Agent Attention, the router grid for mixture-of-top-$k$ attention, and $M_0/\mathcal{P}/\mathcal{C}$ for AVQ2. \Cref{tab:baselines_all} reports every setting we ran. At $N = 18432$ (\cref{tab:video}) each baseline runs at its largest setting from \cref{tab:video_baselines} --- $8$ blocks per row for MRA, an $8\times8\times8$ pool for Agent Attention, a $4\times8\times8$ router grid for mixture-of-top-$k$ attention --- and AVQ2 at $M_0/\mathcal{P}/\mathcal{C} = 64/16/8$ over $128$-token tiles. MRA's probe row there uses $12$ heads of width $64$ rather than the probe's own $16$ heads of width $48$, since its kernels cannot run at width $48$; dense attention costs $55.4$\,ms at that shape.

Two rows depart from that protocol. Agent Attention has no drop-in entry: its agent parameters are new and untrained at initialisation, so a drop-in evaluation would measure an untrained layer. For the same reason it gets three times the fine-tuning budget of every other row, 30 epochs on ADE20K and 3 on ImageNet-1k. Mixture-of-top-$k$ attention is timed wall-clock rather than inside a CUDA graph: its routing sends a data-dependent number of keys to each expert and reads those lengths back on the host, which a graph cannot capture.

\begin{table}[t]
\caption{Every setting of each baseline that we ran, against the AVQ2 row of \cref{tab:baselines},
repeated here for reference. One run per setting, and times are one attention layer at batch size 4.}
\label{tab:baselines_all}
\begin{center}
\small
\begin{tabular}{llccc}
\toprule
Method & Setting & ms & drop-in & fine-tuned \\
\midrule
\multicolumn{5}{l}{\textit{ADE20K, DPT-Large, $N = 901$ (mIoU)}} \\
MRA                & $b = 2$      & 0.848 & 13.81 & 42.26 \\
                   & $b = 4$      & 1.075 & 19.92 & 44.29 \\
                   & $b = 8$      & 1.566 & 31.55 & 45.88 \\
                   & $b = 16$     & 2.542 & 43.72 & 46.27 \\
Agent Attention    & $5\times5$   & 0.237 & ---   & 41.24 \\
                   & $7\times7$   & 0.375 & ---   & 41.52 \\
                   & $10\times10$ & 0.341 & ---   & 42.47 \\
                   & $15\times15$ & 0.576 & ---   & 44.64 \\
Mixture-of-top-$k$ & $5\times5$   & 1.626 & 13.88 & 43.31 \\
                   & $7\times7$   & 2.162 & 21.38 & 46.17 \\
                   & $10\times10$ & 3.326 & 32.68 & 46.44 \\
                   & $15\times15$ & 9.816 & 39.86 & 46.80 \\
AVQ-Attention-2    & $32/8/8$     & \textbf{0.178} & \textbf{44.17} & 46.71 \\
\midrule
\multicolumn{5}{l}{\textit{ImageNet-1k, ViT-B/8, $N = 785$ (top-1)}} \\
MRA                & $b = 2$      & 0.571 & 63.55 & 81.60 \\
                   & $b = 4$      & 0.722 & 78.54 & 84.07 \\
                   & $b = 8$      & 1.041 & 83.64 & 85.32 \\
                   & $b = 16$     & 1.694 & 85.38 & 85.94 \\
Agent Attention    & $4\times4$   & 0.151 & ---   & 73.99 \\
                   & $7\times7$   & 0.183 & ---   & 77.27 \\
Mixture-of-top-$k$ & $5\times5$   & 1.462 & 44.70 & 81.41 \\
                   & $7\times7$   & 2.139 & 79.95 & 84.76 \\
                   & $9\times9$   & 2.653 & 83.67 & 85.52 \\
AVQ-Attention-2    & $32/8/8$     & \textbf{0.125} & 84.75 & 85.60 \\
\bottomrule
\end{tabular}
\end{center}
\end{table}

\section{Local Attention Alone}
\label{app:local}

\Cref{tab:local} gives the settings and the exact numbers behind \cref{fig:local}: Swin and NATTEN across
window sizes, and our exact-local-only layer at three tile sizes.

Swin's speed is not monotone in its window size, which has to do with grid padding and masking and with
kernel artifacts at these sizes, so its points do not order along the speed axis and \cref{fig:local}
draws only our tilings as curves.

\begin{table}[t]
\caption{The numbers behind \cref{fig:local}: every local-attention setting we train, its quality after fine-tuning, and its speedup over FlashAttention-2.}
\label{tab:local}
\begin{center}
\begin{tabular}{llcccc}
\toprule
& & \multicolumn{2}{c}{ADE20K} & \multicolumn{2}{c}{ImageNet-1k} \\
\cmidrule(lr){3-4}\cmidrule(lr){5-6}
Method & Tile & mIoU & Speedup & top-1 & Speedup \\
\midrule
Swin                     & window $5$  &  37.40  & 2.0$\times$         &  77.50    & 2.4$\times$ \\
                         & window $7$  &  40.76    & 2.9$\times$ & 80.98        & 2.3$\times$ \\
                         & window $9$  &  42.23  & 2.1$\times$ &  82.15  & 1.7$\times$ \\
                         & window $11$ & 43.69 & 2.5$\times$ & 83.12 & 2.1$\times$ \\
                         & window $13$ &  43.86  & 1.6$\times$ &  83.81  & 1.4$\times$ \\
NATTEN                   & kernel $5$         & 39.31 & 2.2$\times$ & 79.71 & 1.7$\times$ \\
                         & kernel $7$         & 41.94 & 2.0$\times$ & 82.46 & 1.7$\times$ \\
                         & kernel $9$         & 43.49 & 1.9$\times$ & 83.82 & 1.5$\times$ \\
                         & kernel $11$        & 44.60 & 1.4$\times$ & 84.45 & 1.1$\times$ \\
\midrule
Exact local only, Gilbert tiles   & 32  & 38.50 & 3.5$\times$ &  77.96 & 2.9$\times$ \\
                              & 64  & 40.35 & 3.4$\times$      & 81.41 & 2.8$\times$ \\
                              & 128 & 43.64 & 2.9$\times$ & 83.34 & 2.3$\times$ \\
Exact local only, optimised tiles & 32  & 38.74 & 3.5$\times$ &  78.67 & 2.9$\times$ \\
                              & 64  & 41.97 & 3.4$\times$      & 81.89 & 2.8$\times$ \\
                              & 128 & 44.16 & 2.9$\times$ & 83.62 & 2.3$\times$ \\
\bottomrule
\end{tabular}
\end{center}
\end{table}

\section{V-JEPA Configurations at 9,216 Tokens}
\label{app:vjepa_configs}

\begin{figure}[h]
\begin{center}
\includegraphics[width=0.88\linewidth]{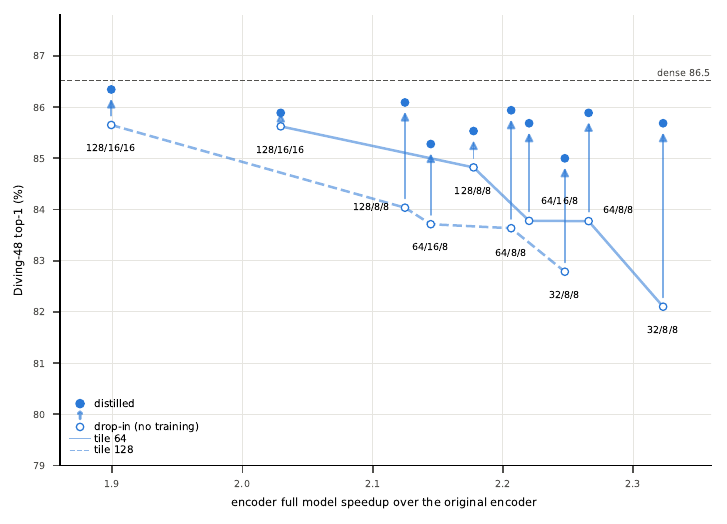}
\end{center}
\caption{Every AVQ-Attention-2 configuration at $N = 9216$, from \cref{fig:vjepa9216}. A label gives the
codebook $M_0/\mathcal{P}/\mathcal{C}$ and the line style the query-tile size; an arrow runs from each
drop-in point to its distilled point.}
\label{fig:vjepa9216_zoom}
\end{figure}

\Cref{fig:vjepa9216_zoom} is the region of \cref{fig:vjepa9216} that the AVQ-Attention-2 configurations occupy, with
each of them named, and \cref{tab:video_baselines} gives the setting behind every baseline point of the
same figure.

Codebook size, refinement budget and tile size all vary across them, and after
distillation the accuracy is remarkably stable: every point lies between $85.00$ and $86.35$ against
dense's $86.51$, over encoder speedups from $1.90\times$ to $2.32\times$. Without distillation the
spread is wider: the smallest codebook, $M_0 = 32$, gives $82.10$--$82.78$ against $85.65$ for
$M_0 = 128$. The larger tile does not pay for itself as a
drop-in, costing speed without much accuracy gain, though the points closest to dense after
distillation do use tile size $128$.

\begin{table}[h]
\caption{Every baseline run on \cref{fig:vjepa9216}: Diving-48 top-1 of the V-JEPA~2 encoder with an
attentive probe at $N = 9216$, one run per setting, with the cost of the encoder beside it.}
\label{tab:video_baselines}
\begin{center}
\small
\begin{tabular}{llcccc}
\toprule
Method & Setting & ms & $\times$ dense & drop-in & distilled \\
\midrule
MRA                & $2$ blocks per row                     & 54.46 & 1.44 & 60.61 & 80.86 \\
MRA                & $4$ blocks per row                     & 60.82 & 1.29 & 66.19 & 82.84 \\
MRA                & $8$ blocks per row                     & 74.14 & 1.05 & 77.16 & 85.18 \\
\midrule
Agent Attention    & $2\times4\times4$ pool, $32$ agents    & 40.25 & 1.94 & ---   & 65.03 \\
Agent Attention    & $2\times8\times8$ pool, $128$ agents   & 42.92 & 1.82 & ---   & 62.59 \\
Agent Attention    & $8\times4\times4$ pool, $128$ agents   & 43.55 & 1.80 & ---   & 70.46 \\
Agent Attention    & $4\times7\times7$ pool, $196$ agents   & 46.28 & 1.69 & ---   & 73.15 \\
Agent Attention    & $4\times8\times8$ pool, $256$ agents   & 47.40 & 1.65 & ---   & 71.62 \\
Agent Attention    & $8\times8\times8$ pool, $512$ agents   & 55.83 & 1.40 & ---   & 77.72 \\
\midrule
Mixture-of-top-$k$ & $2\times4\times4$ pool, $32$ routers   & 50.02 & 1.54 & 20.00 & 55.63 \\
Mixture-of-top-$k$ & $2\times8\times8$ pool, $128$ routers  & 58.53 & 1.31 & 51.52 & 80.51 \\
Mixture-of-top-$k$ & $4\times8\times8$ pool, $256$ routers  & 77.51 & 0.99 & 75.89 & 83.40 \\
\bottomrule
\end{tabular}
\end{center}
\end{table}

\section{Solving for the Tiles}
\label{app:tiling}

\Cref{eq:balanced} is $k$-means with equal-sized clusters, and we solve it the same way. With $\pi$ fixed the centres are weighted means; with $c$ fixed the assignment under the capacity constraint is solved exactly as an assignment problem \citep{bradley2000constrained}, so every tile is full at every iteration rather than only in the limit. We seed the iteration with a recursive bisection of the token set.

Dividing by $\bar a^{(l)}$ in \cref{eq:layerweights} is what keeps $\gamma$ interpretable. It holds the weights at mean $2$ in every layer however large the accumulated displacements have grown, so $\gamma$ sets only how far back the memory reaches and not how strongly the weighting acts. Without it the two would be confounded, and a long memory would also mean a heavier one.

The tiling is solved once, when the model is built, and applied as a permutation folded into the kernels, so it costs nothing per forward pass. One tiling per layer is stored, of one integer per token.

\section{Kernel Pseudocode}
\label{app:kernels}

The two kernels of \cref{sec:method_pipeline}, line by line. \Cref{alg:vqprecompute2} runs once over the keys; \cref{alg:fusedattention2} runs once per query tile and consumes its output.

\begin{algorithm}[h]
\caption{Sorted VQ precompute: one pass over the keys, feeding every query tile of \cref{alg:fusedattention2}.}
\label{alg:vqprecompute2}
\begin{algorithmic}[1]
\Require $K, V \in \mathbb{R}^{BH \times N \times D}$, \; codebook $C \in \mathbb{R}^{H \times M_0(1+\mathcal{C}) \times D}$
\Ensure per code: count $n$, value sum $\bar{V}$, key mean $\bar{K}$
\State $n = 0$; \quad $\bar{V} = 0$; \quad $\bar{K} = 0$
\Statex \textit{// Assign}
\For{each $(bh, \text{Block}_N)$ \textbf{in parallel}} \Comment{$\mathbf{k} = K[bh, \text{Block}_N]$ in registers}
    \State $\mathbf{p} = \arg\min_{m < M_0} \|\mathbf{k} - C[h,m]\|^2$; \quad $\mathbf{c} = \arg\min_{c \,\in\, \text{children}(\mathbf{p})} \|\mathbf{k} - C[h,c]\|^2$ \Comment{$[\text{Block}_N]$ each}
    \State $n[bh, \mathbf{p}] \mathrel{+}= 1$; \quad $n[bh, \mathbf{c}] \mathrel{+}= 1$; \quad store $(\mathbf{p}, \mathbf{c})$ \Comment{Integer atomics}
\EndFor
\Statex \textit{// Counting sort, children of one parent adjacent: every code's keys form one run}
\State $\text{start} = $ exclusive cumsum of $n$ over the children, parent by parent
\For{each $(bh, \text{Block}_N)$ \textbf{in parallel}}
    \State $\text{sorted}[bh,\; \text{atomicAdd}(\text{start}[bh, \mathbf{c}], 1)] = $ key index \Comment{One integer atomic per key}
\EndFor
\Statex \textit{// Reduce, parents and children in one pass}
\For{each $(bh, \text{Block}_N \text{ of sorted})$ \textbf{in parallel}}
    \State $\mathbf{x}_p = \mathbf{y}_p = \mathbf{x}_c = \mathbf{y}_c = 0$ \Comment{$[D]$ each: open parent and child run}
    \For{each key $j$ of the block, in sorted order}
        \State $\mathbf{x}_p, \mathbf{x}_c \mathrel{+}= V[bh, j]$; \quad $\mathbf{y}_p, \mathbf{y}_c \mathrel{+}= K[bh, j]$
        \State \textbf{where} the child run ends at $j$: \; $\bar{V}[bh, c] \mathrel{+}= \mathbf{x}_c$; \quad $\bar{K}[bh, c] \mathrel{+}= \mathbf{y}_c \,/\, n[bh, c]$; \quad $\mathbf{x}_c = \mathbf{y}_c = 0$
        \State \textbf{where} the parent run ends at $j$: \; the same for $p$
    \EndFor
\EndFor
\Statex \textit{// Only a run cut by a block boundary is written by two programs; only there is $\mathrel{+}=$ atomic.}
\end{algorithmic}
\end{algorithm}

\begin{algorithm}[h]
\caption{AVQ attention over cluster means with exact local keys, for one query tile $I$. Refinement runs in rounds of $\text{Block}_{\mathcal{P}}$ parents, $\mathbf{w}$ updated between rounds; the running maximum of the online softmax is left implicit. With codewords instead of means, $\bar K$ is $C$ and the parent logit is rebuilt as $\frac{1}{\mathcal{C}}\sum_c \mathbf{s}_c$.}
\label{alg:fusedattention2}
\begin{algorithmic}[1]
\Require $Q, K, V$; \; $\bar{K}, \bar{V}, n$ and each key's $(p, c)$ from \cref{alg:vqprecompute2}
\Ensure $Y \in \mathbb{R}^{BH \times N \times D}$
\For{each $(bh, I)$ \textbf{in parallel}} \Comment{$\mathbf{q} = Q[bh, I]$ in registers}
    \State $\mathbf{n}^{\text{all}} = n[bh]$; \quad $\mathbf{n} = \mathbf{n}^{\text{all}} - \mathbf{n}^{\,I}$ \Comment{$\mathbf{n}^{\,I}$: histogram of the tile's own $(p, c)$}
    \Statex \quad \textit{// Parents}
    \State $\mathbf{s} = \mathbf{q}\, \bar{K}[bh, 0\!:\!M_0]^\top$; \quad $\mathbf{A} = \exp(\mathbf{s})$; \quad $\mathbf{A}_{:,m} = 0 \;\; \forall m\!: \mathbf{n}^{\text{all}}_m = 0$ \Comment{$[\text{Block}_N, M_0]$}
    \State $\bar{\mathbf{x}} = \mathbf{A}\, \bar{V}[bh, 0\!:\!M_0]$; \quad $\bar{\mathbf{z}} = \textstyle\sum_m \mathbf{A}_{:,m}\, \mathbf{n}_m$; \quad $\mathbf{w}_m = \textstyle\sum_i \mathbf{A}_{im}\, \mathbf{n}_m / \bar{\mathbf{z}}_i$ \Comment{\cref{eq:denomlocal,eq:importance}}
    \Statex \quad \textit{// Refine}
    \For{each $p \in \text{top-}\mathcal{P}(\mathbf{w})$}
        \State $\mathbf{s}_c = \mathbf{q}\, \bar{K}[bh, \text{children}(p)]^\top$ \Comment{$[\text{Block}_N, \mathcal{C}]$}
        \State $\mathbf{s}_p = \textstyle\sum_c (\mathbf{n}^{\text{all}}_c / \mathbf{n}^{\text{all}}_p)\; \mathbf{s}_c$ \Comment{\cref{eq:weighted_parent}}
        \State $\Delta\!\mathbf{A} = \exp(\mathbf{s}_c) - \exp(\mathbf{s}_p)$; \quad $\Delta\!\mathbf{A}_{:,c} = 0 \;\; \forall c\!: \mathbf{n}^{\text{all}}_c = 0$ \Comment{\cref{eq:corr}}
        \State $\bar{\mathbf{x}} \mathrel{+}= \Delta\!\mathbf{A}\; \bar{V}[bh, \text{children}(p)]$; \quad $\bar{\mathbf{z}} \mathrel{+}= \textstyle\sum_c \Delta\!\mathbf{A}_{:,c}\, \mathbf{n}_c$
    \EndFor
    \Statex \quad \textit{// Exact local keys}
    \State $\mathbf{k}, \mathbf{v} = K[bh, I],\, V[bh, I]$; \quad $m(j) = $ key $j$'s $c$ if its $p$ was refined, else its $p$
    \State $\mathbf{A}^{\text{loc}} = \exp(\mathbf{q}\, \mathbf{k}^\top)$; \quad $\bar{\mathbf{x}} \mathrel{+}= \big(\mathbf{A}^{\text{loc}} - \exp(\mathbf{q}\, \bar{K}[bh, m(\cdot)]^\top)\big)\, \mathbf{v}$; \quad $\bar{\mathbf{z}} \mathrel{+}= \textstyle\sum_j \mathbf{A}^{\text{loc}}_{:,j}$ \Comment{\cref{eq:localexact}}
    \State $Y[bh, I] = \bar{\mathbf{x}} \,/\, \bar{\mathbf{z}}$
\EndFor
\end{algorithmic}
\end{algorithm}

\section{Backward Pass for Cluster Means}
\label{app:derivations}

Write $\ell_{im} = Q_i \bar K_m^{\top}$ for the logit of cluster $m$ at query $i$, $p_{im} = \exp(\ell_{im})\, n_m / \sum_{m'} \exp(\ell_{im'})\, n_{m'}$ for its weight, and
\begin{equation}
\label{eq:app_forward}
O_i \;=\; \sum_{m} p_{im}\, \bar V_m / n_m
\end{equation}
for the layer output, which is \cref{eq:vqattention} with $\bar K_m$ in place of the codeword. Let $dO_i = \partial L / \partial O_i$. Three steps give \cref{eq:collapse}.

\paragraph{Through the softmax.}
$p_{im}$ is a softmax over $\ell_{im}$ with fixed positive weights $n_m$, so $\partial p_{im'} / \partial \ell_{im} = p_{im'}(\mathbf{1}[m'{=}m] - p_{im})$ as for a plain softmax. Differentiating \cref{eq:app_forward} and using $\sum_{m'} p_{im'} \bar V_{m'}/n_{m'} = O_i$,
\begin{equation}
\label{eq:app_softmax}
\frac{\partial L}{\partial \ell_{im}}
\;=\; p_{im}\Big( \big\langle dO_i, \bar V_m/n_m \big\rangle - \sum_{m'} p_{im'} \big\langle dO_i, \bar V_{m'}/n_{m'} \big\rangle \Big)
\;=\; p_{im}\, \big\langle dO_i,\ \bar V_m/n_m - O_i \big\rangle .
\end{equation}
The bracket is where the per-key value $V_j$ disappears: it enters only through $\bar V_m = \sum_{j \in m} V_j$, so a sum over $N$ keys has already become one over $M$ clusters.

\paragraph{Through the logit.}
$\partial \ell_{im} / \partial \bar K_{m,d} = Q_{i,d}$, and $n_m$ does not depend on $\bar K_m$. Summing \cref{eq:app_softmax} over queries,
\begin{equation}
\label{eq:app_kbar}
\frac{\partial L}{\partial \bar K_{m,d}} \;=\; \sum_i p_{im}\, \big\langle dO_i,\ \bar V_m/n_m - O_i \big\rangle\, Q_{i,d} .
\end{equation}

\paragraph{Through the mean.}
The assignment $m(\cdot)$ is piecewise constant, so neither it nor $n_m$ carries a gradient, and the only path from $\bar K_m$ back to a key is the mean itself: $\partial \bar K_{m,d}/\partial K_{j',d} = \mathbf{1}[m(j') = m]/n_m$. Composing with \cref{eq:app_kbar} gives \cref{eq:collapse}.

\paragraph{Check.}
Summing \cref{eq:collapse} over the $n_m$ keys of a cluster returns \cref{eq:app_kbar} exactly: the cluster's gradient is shared equally among its keys, and none of it is created or lost. Every factor on the right of \cref{eq:collapse} is either per cluster ($\bar V_m$, $n_m$) or per query ($p_{im}$, $O_i$, $dO_i$, $Q_i$), and $p_{im}$ is already formed in the forward pass, so the backward costs $\mathcal{O}(MND)$ like the forward. No $N \times N$ matrix is built at any point.

\section{Skipping the Double-Counting Correction}
\label{app:count_correction}

A tile's local keys are attended to at their own logits by \cref{eq:localexact} and are also members of a
cluster, so without the discount of \cref{sec:method_localexact} each contributes twice, once exactly and
once through its cluster's mean key. The second copy inflates that key's weight, while the larger
normaliser dilutes every key outside the tile.

\begin{table}[h]
\caption{Drop-in accuracy of every configuration of \cref{fig:cost}, with and without the correction of
\cref{sec:method_localexact}. ImageNet is top-1, ADE20K is mean IoU.}
\label{tab:double_counting}
\begin{center}
\small
\begin{tabular}{llccc}
\toprule
& $M_0/\mathcal{P}/\mathcal{C}$ & drop-in & no double-counting correction & $\Delta$ \\
\midrule
ImageNet & $16/8/4$   & $82.92$ & $82.40$ & $-0.52$ \\
         & $32/4/4$   & $83.59$ & $82.88$ & $-0.71$ \\
         & $32/8/4$   & $84.05$ & $83.51$ & $-0.54$ \\
         & $32/12/4$  & $84.38$ & $83.63$ & $-0.75$ \\
         & $32/4/8$   & $84.23$ & $83.65$ & $-0.58$ \\
         & $32/8/8$   & $84.75$ & $84.24$ & $-0.51$ \\
         & $32/12/8$  & $84.80$ & $84.33$ & $-0.47$ \\
\midrule
ADE20K   & $16/8/4$   & $43.03$ & $41.90$ & $-1.13$ \\
         & $32/4/4$   & $43.36$ & $43.13$ & $-0.23$ \\
         & $32/8/4$   & $43.79$ & $43.42$ & $-0.37$ \\
         & $32/12/4$  & $43.92$ & $43.01$ & $-0.91$ \\
         & $32/4/8$   & $43.97$ & $42.94$ & $-1.03$ \\
         & $32/8/8$   & $44.17$ & $43.92$ & $-0.25$ \\
         & $32/12/8$  & $44.73$ & $43.36$ & $-1.37$ \\
         & $64/8/8$   & $45.26$ & $43.89$ & $-1.37$ \\
\bottomrule
\end{tabular}
\end{center}
\end{table}

Without the correction the accuracy is lower in all configurations of \cref{tab:double_counting}.

\section{Analysis of the Parent Refinement Criterion}
\label{app:refinement_score}
\label{app:refine_criterion}

Refinement spends a fixed budget of $\mathcal{P}$ parents per query tile, and \cref{eq:importance} spends it on the parents that receive the most attention mass. Attention mass, however, is a property of where a query looks, not of how badly a cluster is represented by its mean: a cluster whose keys are identical is approximated exactly, however much mass it holds. Within-cluster spread is the obvious alternative. As elsewhere, the softmax scale $1/\sqrt{D}$ is absorbed into $Q$.

\subsection{The error of a cluster-mean approximation}
\label{app:refinement_error}

Fix a query $Q_i$. Every key is assigned to one cluster by $m(\cdot)$, so cluster $m$ holds the $n_m$ keys $\{\, j : m(j) = m \,\}$, with mean key $\bar K_m$ as in \cref{eq:keymeans} and mean value $\bar V_m / n_m$ ($\bar V_m$ is a sum, $\bar K_m$ a mean). Averages over the keys of one cluster are written $\langle f \rangle_m = \frac{1}{n_m} \sum_{j:\, m(j) = m} f_j$. Write $Y_i$ for the output of exact attention, $Y_i = \sum_j \exp(Q_i K_j^{\top}) V_j / \sum_l \exp(Q_i K_l^{\top})$, and $O_i = \sum_m p_{im}\, \bar V_m / n_m$ for the cluster-mean approximation of it, where $p_{im} = \exp(Q_i \bar K_m^{\top})\, n_m / \bar Z_i$ is the cluster's share of the attention and $\bar Z_i = \sum_{m'} \exp(Q_i \bar K_{m'}^{\top})\, n_{m'}$, as in \cref{eq:vqattention}. Write
\begin{equation}
\label{eq:app_u}
u_j \;=\; Q_i \big(K_j - \bar K_{m(j)}\big)^{\!\top} ,
\end{equation}
the logit each key has in excess of its cluster's, so that $\langle u \rangle_m = 0$ within every cluster by construction. Each cluster's exact contribution to the denominator of $Y_i$, and then to its numerator, factorises as
\begin{equation}
\label{eq:app_cluster_exact}
\begin{aligned}
\sum_{j:\, m(j) = m} \exp(Q_i K_j^{\top}) \;&=\; n_m \exp(Q_i \bar K_m^{\top})\,\big\langle e^{u} \big\rangle_m , \\[2pt]
\sum_{j:\, m(j) = m} \exp(Q_i K_j^{\top})\, V_j \;&=\; n_m \exp(Q_i \bar K_m^{\top})\,\big\langle e^{u} V \big\rangle_m .
\end{aligned}
\end{equation}
The approximation replaces both averages by their zeroth-order values, $1$ and $\bar V_m / n_m$. Expanding the exponential and averaging term by term,
\begin{equation}
\label{eq:app_moments}
\begin{aligned}
\big\langle e^{u} \big\rangle_m
  &\;=\; \big\langle 1 + u + \tfrac{1}{2} u^2 + \cdots \big\rangle_m
  \;\approx\; 1 + \tfrac{1}{2} \big\langle u^2 \big\rangle_m
  \;=\; 1 + \alpha_{im} , \\[4pt]
\big\langle e^{u} V \big\rangle_m
  &\;=\; \Big\langle \big( 1 + u + \tfrac{1}{2} u^2 + \cdots \big)
         \Big( \tfrac{\bar V_m}{n_m} + \Delta V \Big) \Big\rangle_m \\[2pt]
  &\;\approx\; \tfrac{\bar V_m}{n_m} \Big( 1 + \tfrac{1}{2} \big\langle u^2 \big\rangle_m \Big)
         + \big\langle u\, \Delta V \big\rangle_m
  \;=\; \frac{\bar V_m}{n_m} \big( 1 + \alpha_{im} \big) + \varepsilon_{im} ,
\end{aligned}
\end{equation}
where every term linear in $u$ or in $\Delta V$ alone drops out, since $\langle u \rangle_m = 0$ by \cref{eq:app_u} and $\langle \Delta V \rangle_m = 0$ by definition of the mean. The two surviving moments are
\begin{equation}
\label{eq:app_alpha_eps}
\begin{aligned}
\alpha_{im} \;&=\; \tfrac{1}{2} \big\langle u^2 \big\rangle_m \;=\; \tfrac{1}{2}\, Q_i \Sigma_m Q_i^{\top} ,
&\qquad \Sigma_m \;&=\; \big\langle \Delta K^{\top} \Delta K \big\rangle_m , \\[2pt]
\varepsilon_{im} \;&=\; \big\langle u\, \Delta V \big\rangle_m \;=\; Q_i \Gamma_m ,
&\qquad \Gamma_m \;&=\; \big\langle \Delta K^{\top} \Delta V \big\rangle_m ,
\end{aligned}
\end{equation}
with $\Delta K = K_j - \bar K_m$ and $\Delta V = V_j - \bar V_m / n_m$, the second equality in each line following from $u = Q_i \Delta K^{\top}$, so $\Sigma_m$ is the within-cluster covariance of the \emph{keys} and $\Gamma_m$ the within-cluster \emph{key--value cross-covariance}. Summing \cref{eq:app_cluster_exact} over the clusters and substituting \cref{eq:app_moments},
\begin{equation}
\label{eq:app_ratio}
\begin{aligned}
Y_i \;&=\; \frac{\sum_m n_m \exp(Q_i \bar K_m^{\top}) \big[\, (\bar V_m / n_m)(1 + \alpha_{im}) + \varepsilon_{im} \,\big]}
              {\sum_m n_m \exp(Q_i \bar K_m^{\top})\, (1 + \alpha_{im})} \\[6pt]
    \;&=\; \frac{\sum_m p_{im} \big[\, (\bar V_m / n_m)(1 + \alpha_{im}) + \varepsilon_{im} \,\big]}
              {\sum_m p_{im}\, (1 + \alpha_{im})} \\[6pt]
    \;&=\; \frac{O_i + \sum_m p_{im} \big[\, \alpha_{im}\, \bar V_m / n_m + \varepsilon_{im} \,\big]}
              {1 + \sum_m p_{im}\, \alpha_{im}} ,
\end{aligned}
\end{equation}
using $n_m \exp(Q_i \bar K_m^{\top}) / \bar Z_i = p_{im}$, $\sum_m p_{im} = 1$ and $\sum_m p_{im}\, \bar V_m / n_m = O_i$. Writing $A_i = \sum_m p_{im}\, \alpha_{im}$ and $B_i = \sum_m p_{im}\big[\, \alpha_{im}\, \bar V_m / n_m + \varepsilon_{im} \,\big]$ for the two corrections, the denominator expands as $1/(1 + A_i) = 1 - A_i + A_i^2 - \cdots$, so
\begin{equation}
\label{eq:app_expand}
Y_i \;=\; \frac{O_i + B_i}{1 + A_i} \;=\; (O_i + B_i)\,\big( 1 - A_i + A_i^2 - \cdots \big)
\;=\; O_i + B_i - A_i O_i + \mathcal{O}(u^3) .
\end{equation}
$\varepsilon_{im}$ is linear in $u$ and $\alpha_{im}$ quadratic, so the discarded $A_i B_i$ and $A_i^2 O_i$ are of the order already dropped in \cref{eq:app_moments}. Rearranging $O_i - Y_i = A_i O_i - B_i$ gives
\begin{equation}
\label{eq:app_error}
\boxed{\;
O_i - Y_i \;=\; -\sum_{m} p_{im} \Big[\, \alpha_{im} \big( \bar V_m / n_m - O_i \big) \;+\; \varepsilon_{im} \,\Big] \; }
\end{equation}

From \cref{eq:app_error} the attention mass $p_{im}$ is a direct factor of the error caused by clustering keys, not merely correlated with it, so the question is not whether to rank by mass or by something else, such as per-cluster variance, but what to multiply the mass with. The bracket holds two distinct failures. $\alpha_{im}$ is \emph{mis-weighting}: within-cluster key spread inflates the cluster's share of the softmax, since $\langle e^u\rangle_m \geq 1$ by Jensen, and the cluster is counted too heavily. The cost of counting it too heavily is proportional to how far its value mean lies from the output, $\bar V_m / n_m - O_i$, so a high-variance cluster whose value mean coincides with the output costs nothing. $\varepsilon_{im}$ is \emph{mis-valuing}: even when the cluster's total weight is right, exact attention does not weight the cluster's own keys equally, but gives more to the ones scoring above its mean. If those keys carry different values from the ones scoring below, the correct contribution is a weighted average of its values where $\bar V_m / n_m$ is the unweighted one. The key variance enters only $\alpha_{im}$; $\varepsilon_{im}$ comes from the key--value cross-covariance, which the key variance bounds but does not determine.

The above does not assume nor does it break under exact local attention: a query tile's own keys are attended to at their true logits by \cref{eq:localexact}, so they contribute no error, and the derivation applies to each cluster's remaining keys with $n_m$, $\bar V_m$ and $\langle \cdot \rangle_m$ restricted to the keys outside the tile. One detail differs: the key mean $\bar K_m$ is deliberately not corrected per tile, so $\langle u \rangle_m$ no longer vanishes but equals $\mu_{im} = Q_i (\bar K_m^{\,\setminus I} - \bar K_m)^{\top}$, which adds a first-order term proportional to the share of a cluster's keys inside the tile.

\subsection{From the error to a score}
\label{app:refinement_score_form}

\cref{eq:app_error} is one term per cluster. Write $E_{im}$ for cluster $m$'s,
\begin{equation}
\label{eq:app_term}
E_{im} \;=\; -\,p_{im} \Big[\, \alpha_{im} \big( \bar V_m / n_m - O_i \big) \;+\; \varepsilon_{im} \,\Big] ,
\qquad O_i - Y_i \;=\; \sum_m E_{im} .
\end{equation}
Refining parent $m$ replaces $E_{im}$ by the terms of its children, which changes the error by
\begin{equation}
\label{eq:app_gain_def}
g_{im} \;=\; \sum_c E_{ic} \;-\; E_{im} .
\end{equation}
Writing the children's masses as $p_{ic}$,
\begin{equation}
\label{eq:app_cancel}
\begin{aligned}
\frac{g_{im}}{p_{im}} \;&=\; \Big[\, \alpha_{im} \Big( \frac{\bar V_m}{n_m} - O_i \Big) + \varepsilon_{im} \,\Big]
\;-\; \sum_c \frac{p_{ic}}{p_{im}} \Big[\, \alpha_{ic} \Big( \frac{\bar V_c}{n_c} - O_i \Big) + \varepsilon_{ic} \,\Big] \\[2pt]
&=\; \Big[\, \alpha_{im} \Big( \frac{\bar V_m}{n_m} - O_i \Big) + \varepsilon_{im} \,\Big]
\;-\; \sum_c \frac{n_c}{n_m}\, e^{Q_i (\bar K_c - \bar K_m)^{\top}} \Big[\, \alpha_{ic} \Big( \frac{\bar V_c}{n_c} - O_i \Big) + \varepsilon_{ic} \,\Big] .
\end{aligned}
\end{equation}
The normaliser cancels in $p_{ic}/p_{im}$, leaving the count ratio and the shift between the child's mean key and the parent's. That exponential multiplies $\alpha_{ic}$ and $\varepsilon_{ic}$, which are second order already, so expanding it beyond its leading term contributes at third order. This leaves
\begin{equation}
\label{eq:app_gain_trunc}
\frac{g_{im}}{p_{im}} \;\approx\; \Big[\, \alpha_{im} \Big( \frac{\bar V_m}{n_m} - O_i \Big) + \varepsilon_{im} \,\Big]
\;-\; \sum_c \frac{n_c}{n_m} \Big[\, \alpha_{ic} \Big( \frac{\bar V_c}{n_c} - O_i \Big) + \varepsilon_{ic} \,\Big] .
\end{equation}

Because every key descends to a child (\cref{eq:weighted_parent}), the parent's keys are partitioned among them: $n_m = \sum_c n_c$, $\bar V_m = \sum_c \bar V_c$ and $\bar K_m = \sum_c (n_c/n_m) \bar K_c$. Each key of the parent lies in exactly one child, so its deviation from the parent mean splits into a within-child step and a child-to-parent one,
\begin{equation}
\label{eq:app_split}
K_j - \bar K_m \;=\; \big( K_j - \bar K_{c(j)} \big) \;+\; \big( \bar K_{c(j)} - \bar K_m \big) ,
\end{equation}
and in the second moment the cross term between the two averages to zero, since $\langle K_j - \bar K_c \rangle_c = 0$ within each child while $\bar K_c - \bar K_m$ is constant over it. Both statistics of \cref{eq:app_alpha_eps} therefore split over the partition,
\begin{equation}
\label{eq:app_decomp}
\alpha_{im} \;=\; \alpha^{\mathrm{b}}_{im} + \sum_c \frac{n_c}{n_m}\, \alpha_{ic} ,
\qquad
\varepsilon_{im} \;=\; \varepsilon^{\mathrm{b}}_{im} + \sum_c \frac{n_c}{n_m}\, \varepsilon_{ic} ,
\end{equation}
into a between-child part, built from the child means alone,
\begin{equation}
\label{eq:app_between}
\begin{aligned}
\Sigma^{\mathrm{b}}_m \;&=\; \sum_c \frac{n_c}{n_m} \big( \bar K_c - \bar K_m \big)^{\top}\big( \bar K_c - \bar K_m \big) , \\[2pt]
\Gamma^{\mathrm{b}}_m \;&=\; \sum_c \frac{n_c}{n_m} \big( \bar K_c - \bar K_m \big)^{\top}\Big( \frac{\bar V_c}{n_c} - \frac{\bar V_m}{n_m} \Big) ,
\end{aligned}
\end{equation}
and a count-weighted average of the children's own. \Cref{eq:weighted_parent} puts the reference point at the parent's own mean key, so $\Sigma^{\mathrm{b}}_m$ is the covariance of the child means and vanishes when they coincide, and $\langle u \rangle_m = 0$ at the parent as well as at each child.

Substituting \cref{eq:app_decomp} into \cref{eq:app_gain_trunc}, the children's $\varepsilon_{ic}$ cancel against the within-child part of $\varepsilon_{im}$, and their $\alpha_{ic}$ against the within-child part of $\alpha_{im}$:
\begin{equation}
\label{eq:app_cancelled}
\frac{g_{im}}{p_{im}} \;\approx\; \alpha^{\mathrm{b}}_{im} \Big( \frac{\bar V_m}{n_m} - O_i \Big) \;+\; \varepsilon^{\mathrm{b}}_{im} \;+\; \sum_c \frac{n_c}{n_m}\, \alpha_{ic} \Big( \frac{\bar V_m}{n_m} - \frac{\bar V_c}{n_c} \Big) .
\end{equation}
The remainder is a within-child key variance times a between-child value contrast, third order again, so
\begin{equation}
\label{eq:app_gain}
g_{im} \;\approx\; p_{im} \Big[\, \alpha^{\mathrm{b}}_{im} \Big( \frac{\bar V_m}{n_m} - O_i \Big)
\;+\; \varepsilon^{\mathrm{b}}_{im} \,\Big] ,
\end{equation}
to the order kept in \cref{eq:app_error}. Refinement removes the between-child part of both statistics and leaves the within-child part: a parent whose children coincide has $\Sigma^{\mathrm{b}}_m = \Gamma^{\mathrm{b}}_m = 0$ and gains nothing from being refined, however far apart its keys are. The prefactor in \cref{eq:app_gain} is the same $p_{im}$ as in \cref{eq:app_error}. Attention mass is therefore proportional not only to the error a cluster causes, but to the error reduction obtained by replacing its mean with its children's means. The per-tile score it implies is the size of that gain over the tile's queries,
\begin{equation}
\label{eq:app_score}
\sigma_m(I) \;=\; \sum_{i \in I} \big\lVert g_{im} \big\rVert ,
\end{equation}
and \cref{eq:importance} is \cref{eq:app_score} with the bracket of \cref{eq:app_gain} replaced by a constant: $w_m(I) = \sum_{i \in I} p_{im}$.

The per-cluster errors in \cref{eq:app_error} are vectors. They add, and they can cancel, so the $\mathcal{P}$ parents whose refinement helps most jointly are in general not the $\mathcal{P}$ with the largest individual gains. Any criterion that scores each parent independently and takes the top $\mathcal{P}$ --- including \cref{eq:app_score} --- is therefore an approximation. We quantify the gap in \cref{app:refinement_results} by comparing against a selection that re-scores after each choice.

\subsection{Measured error reduction}
\label{app:refinement_results}
\label{app:refine_ablation}

We report the relative attention-output error $\lVert O - Y \rVert / \lVert Y \rVert$ per layer on the drop-in configurations of \cref{sec:experiments}, varying how many parents are refined and which criterion selects them. We measure that refining a quarter of the parents by attention mass (\cref{eq:importance}) --- the budget these configurations run, $\mathcal{P}/M_0 = 8/32$, $8/32$ and $16/64$ --- recovers $76\%$, $85\%$ and $92\%$ of the error reduction that refining all of them gives. However, ranking by attention mass keeps only the prefactor of \cref{eq:app_score}, so it is a proxy for the ideal score.

Below we compare it against other candidate criteria. Each criterion is measured against the best selection we can identify: refine one parent at a time, each time taking the parent that reduces the measured error most. \Cref{tab:refinement_criteria} reports the share of that selection's error reduction each criterion achieves.

\begin{table}[h]
\caption{Share of the error reduction the sequential reference reaches at the same budget $\mathcal{P}$, per selection criterion, averaged over layers: $1$ matches it, $0$ is no better than refining nothing. Drop-in; $M_0/\mathcal{P}/\mathcal{C} = 32/8/8$ on the two image tasks and $64/16/8$ at $N = 9216$.}
\label{tab:refinement_criteria}
\begin{center}
\small
\begin{tabular}{lccc}
\toprule
criterion for the top-$\mathcal{P}$ & ImageNet & ADE20K & V-JEPA~2 \\
\midrule
random & $0.247$ & $0.229$ & $0.279$ \\
key variance & $0.486$ & $0.613$ & $0.331$ \\
attention mass, \cref{eq:importance} (ours) & $0.789$ & $0.881$ & $0.918$ \\
mass $\times$ key variance & $0.815$ & $0.910$ & $0.921$ \\
\midrule
top-$\mathcal{P}$ by each parent's own reduction & $0.981$ & $0.981$ & $0.983$ \\
sequential reference & $1.000$ & $1.000$ & $1.000$ \\
\bottomrule
\end{tabular}
\end{center}
\end{table}

Attention mass alone as a criterion already reaches most of the reference's reduction, and most of all where the sequence is longest: $0.918$ at $N = 9216$. The variance here is the trace of the within-cluster key covariance, the form the VQ-precompute kernel can carry as one scalar per parent. Multiplying mass by the key variance improves a little --- by \cref{eq:app_decomp} refinement removes only the part of that variance the children's means account for, so the statistic still differs from the quantity the gain depends on. At up to $7\%$ of layer time for at most $0.03$ more of the reference's reduction, we decided not to use it. Proxies closer to the derived form get more expensive still: the directional $Q_i \Sigma_m Q_i^{\top}$ of \cref{eq:app_alpha_eps} would need a $D \times D$ matrix per parent and a quadratic form at every query.

\end{document}